# Redundancy in Logic II: 2CNF and Horn Propositional Formulae

Paolo Liberatore*

**Abstract**

We report complexity results about redundancy of formulae in 2CNF form. We first consider the problem of checking redundancy and show some algorithms that are slightly better than the trivial one. We then analyze problems related to finding irredundant equivalent subsets (I.E.S.) of a given set. The concept of cyclicity proved to be relevant to the complexity of these problems. Some results about Horn formulae are also shown.

## 1 Introduction

The complexity of some problems related to the redundancy of propositional CNF formulae has been studied in a previous paper [Lib05]. The motivations for studying redundancy can be summarized as follows: first, removing redundancy from a knowlegde base makes it simpler without changing much its structure; second, the presence of redundant parts in a knowledge base can be a sign of importance of the represented concept, but can also be a sign of mistakes in the formulation of the knowledge base [Lib05]. Related to the problem of redundancy of CNF formulae are the redundancy of production rules [Gin88, SS97], the minimiziation of CNF and Horn formulae [MS72, Mai80, ADS86, HK93, HW97, Uma98], the redundancy of literals in a clause [GF93], the redundancy for non-monotonic logics [Lib], the equivalence and extension-equivalence of irredundant formulae [BZ05], and the problem of minimal unsatisfiability [PW88, FKS02, Bru03], which is the special case of irredundancy of inconsistent formulae. A comparison between the problem of redundancy and related work is given in the paper where redundancy of general CNF formulae is studied [Lib05].

In this paper, we study the complexity of problems related to the redundancy of formulae in 2CNF and Horn form. Most of the results are about the 2CNF form, as the corresponding problems for the Horn case are either trivial or have proofs of complexity that coincide with the corresponding ones for the 2CNF form.

The first problem we consider is that of checking the redundancy of a 2CNF formula. Since a formula is redundant if and only if it is equivalent to one of its subsets and checking equivalence for 2CNF formulae is polynomial, the problem is polynomial. We slightly improve over the trivial algorithm by showing that redundancy can be checked in time $O(nm)$, where $n$ is the number of variables and $m$ is the number of clauses of the formula.

*Dipartimento di Informatica e Sistemistica, Università di Roma "La Sapienza", Via Salaria 113, 00198 Roma, Italy. Email: `paolo@liberatore.org`

The other problems we consider are about the irredundant equivalent subsets (i.e.s.) of a formula. In particular, the following problems are easily shown to be polynomial for formulae in 2CNF: check whether a formula is an i.e.s. of another one; check whether a clause is in all i.e.s.'s of a formula; and check whether a formula has an unique i.e.s.. The last two problems are polynomial thanks to the following results [Lib05]: a clause $\gamma$ is in all i.e.s.'s of a formula $\Pi$ if and only if $\Pi \backslash \{\gamma\} \models \gamma$; a formula $\Pi$ has an unique i.e.s. if and only if $\{\gamma \in \Pi \mid \Pi \backslash \{\gamma\} \not\models \gamma\} \models \Pi$. Combined with the fact that inference is polynomial for 2CNF clauses, these two results imply that checking the presence of a clause in an i.e.s. and the uniqueness of i.e.s.'s are polynomial problems.

Two problems about i.e.s.'s require a more complicated complexity analysis: checking whether a clause is in at least an i.e.s. of a formula and checking whether a formula has an i.e.s. of size bounded by an integer $k$. The complexity of these two problems largely depend on the presence of cycles of clauses in the formula. Namely, if the formula contains a cycle of clauses, defined as a sequence of clauses $[\neg l_1 \vee l_2, \neg l_2 \vee l_3, \ldots, \neg l_n \vee l_1]$, these two problems are typically NP-complete, while they are polynomial if the formula does not contain cycles.

The complexity analysis for the 2CNF form has been carried on separately for the cases in which the set of clauses is:

1. inconsistent;

2. consistent and implying some literals;

3. consistent and not implying any literal.

We prove that the clauses not containing implied literals and the clauses containing implied literals can be considered separately. More precisely, the problem of redundancy can be solved in two steps:

1. check the redundancy in $\Pi$ of clauses $l \vee l'$ such that $\Pi \models l$;

2. remove from $\Pi$ all clauses $l \vee l'$ such that $\Pi \models l$, and check redundancy.

A similar procedure can be used for problems about i.e.s.'s: we indeed prove that every i.e.s. of a consistent formula is composed of two parts, the second being an i.e.s. of the formula composed of the clauses of the formula not containing an implied literal. An i.e.s. of a formula can therefore be found by first finding an i.e.s. of this reduced formula and then checking which clauses have to be added to allow the derivation of literals that are implied by the original formula.

The three conditions of inconsistent formulae, formulae implying literals, and formulae implying literals, require each a different analysis. Surprisingly, however, the complexity of the problems is usually the same in the three cases. Namely, the complexity of checking redundancy is always $O(nm)$ regardless of these conditions, while the complexity of the problems of presence in an i.e.s. and of the uniqueness of i.e.s.'s depend more on the presence of cycles in the formula than on the consistency or presence of implied literals.

# 2 Preliminaries

In this paper, we study CNF formulae that are either in 2CNF or in Horn form. Given a set of propositional variables, a literal is a variable or a negated variable. A clause is a disjunction of literals; in particular, a unary/binary clause is a clause composed of one or two literals. An Horn clause is a clause containing at most one positive literal. A set of clauses containing only unary and binary clauses is a 2CNF formula. A set of clauses composed only of Horn clauses is an Horn formula.

Redundancy of clauses and formulae are defined as follows.

**Definition 1** *A clause $\gamma$ is redundant in a CNF formula $\Pi$ if $\Pi \backslash \{\gamma\} \models \gamma$.*

This definition allows a clause $\gamma$ not in $\Pi$ to be classified as redundant in $\Pi$. However, we are typically interested into the redundancy of clauses $\gamma \in \Pi$. Obviously, if $\gamma \in \Pi$ is irredundant in $\Pi$, it is also irredundant in every $\Pi' \subseteq \Pi$.

**Definition 2** *A CNF formula is redundant if it contains a redundant clause.*

In this paper we study the problem of checking the redundancy of 2CNF and Horn formulae, and some problems related to making a formula irredundant by eliminating redundant clauses. What results from this process is formalized by the following definition.

**Definition 3 ([Lib05])** *An Irredundant Equivalent Subset (I.E.S.) of a CNF formula $\Pi$ is a formula $\Pi'$ such that $\Pi' \subseteq \Pi$, $\Pi' \equiv \Pi$, and $\Pi'$ is irredundant.*

Every formula has at least one I.E.S. An irredundant formula has a single I.E.S., which is the formula itself. A redundant I.E.S. can have a number of I.E.S.'s ranging from one to exponentially many [Lib05]. The following properties have been proved in a previous paper.

**Property 1 ([Lib05])** *A clause $\gamma$ is in all I.E.S.'s of a formula $\Pi$ if and only if $\gamma$ is irredundant in $\Pi$.*

**Property 2 ([Lib05])** *A formula $\Pi$ has a single I.E.S. if and only if $\{\gamma \in \Pi \mid \Pi \backslash \{\gamma\} \not\models \gamma\}\} \models \Pi$.*

We use the following notation for the literals that are entailed by a formula.

NOTATION: $$\Pi_{\models} = \{l \mid \Pi \models l\}$$

The following notation for the clauses containing a literal is a set will be used.

NOTATION: $$\Pi | \{l_1, \ldots, l_m\} = \{\gamma \mid l_i \in \gamma,\ \gamma \in \Pi\}$$

We also use $\Pi | l = \Pi | \{l\}$, where $l$ is a literal.

## 2.1 Unit Propagation

The proofs of complexity of redundancy of 2CNF formulae are mostly done using unit propagation. The following lemmas show how entailment is related to unit propagation. We denote by $\models_R$ the derivation by resolution, and by $\models_{UP}$ the derivation by unit propagation.

In what follows, $\Pi$ denotes a 2CNF formula, i.e., a set of clauses, each composed at most of two literals. We assume that $\Pi$ does not contain the empty clause. By resolution trees we mean regular resolution trees; their root is labeled with a clause which is not necessarily $\bot$. A well-known property of resolution is that of being a complete inference method for prime implicates:

**Property 3** *For any set of clauses $\Pi$ and clause $\gamma$, it holds $\Pi \models \gamma$ if and only if there exists $\gamma' \subseteq \gamma$ such that $\Pi \models_R \gamma'$.*

When applied to binary clauses, this property can be reformulated as:

$\Pi \models l_1 \vee l_2$ if and only if one of the following conditions hold:  $\Pi \models_R \bot$
$\Pi \models_R l_1$
$\Pi \models_R l_2$
$\Pi \models_R l_1 \vee l_2$

We show how resolution is related to unit propagation for 2CNF formulae.

**Lemma 1** *For any 2CNF formula $\Pi$ and two literals $l_1$ and $l_2$, if $\Pi \models_R l_1 \vee l_2$, then $\Pi \cup \{\neg l_1\} \models_{UP} l_2$.*

*Proof.* Since $\Pi \models_R l_1 \vee l_2$ there is a resolution tree rooted with $l_1 \vee l_2$. We prove the lemma by induction on the height of the tree. The base case of recursion is when the tree is a leaf. In this case, $l_1 \vee l_2 \in \Pi$, which implies that $\Pi \cup \{\neg l_1\} \models_{UP} l_2$.

Let us therefore assume that the claim holds for any binary clause that can be proved with a tree of height $k$, and prove it for clauses requiring trees of height $k+1$. Let us therefore consider a tree of height $k + 1$ and labeled with $l_1 \vee l_2$ in the root. Its subtrees have height less than or equal to $k$, and their roots are marked with $l_1 \vee l_3$ and $\neg l_3 \vee l_2$ for some literal $l_3$; note that resolution does not allow to derive $l_1 \vee l_2$ from $l_1$ or from $l_2$.

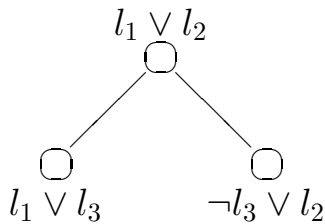

Since the resolution trees of $l_1 \vee l_3$ and $\neg l_3 \vee l_2$ have both height less than or equal to $k$, by the induction hypothesis both $\Pi \cup \{\neg l_1\} \models_{UP} l_3$ and $\Pi \cup \{l_3\} \models_{UP} l_2$ hold. As a result, $\Pi \cup \{\neg l_1\} \models_{UP} l_2$.  □

The following lemma shows that inference over a single unit clause can be checked using unit propagation.

**Lemma 2** *For any 2CNF formula $\Pi$ and literal $l$, if $\Pi \models_R l$ then $\Pi \cup \{\neg l\} \models_{UP} \bot$.*

*Proof.* The claim is proved by induction on the height of the resolution tree rooted with $l$. In the base case, this tree is a leaf, and therefore $l \in \Pi$. The claim $\Pi \cup \{\neg l\} \models_{UP} \bot$ therefore holds.

We now assume that the claim is true for any literal that is derivable from $\Pi$ using a resolution tree of height less than or equal to $k$ and prove that the same holds for height $k+1$. Let $l$ be the root of a resolution tree of of height $k+1$. The root of this tree is $l$, and its children can be either both binary, or a unary clause resolved with a binary clause. Let us consider this latter case first.

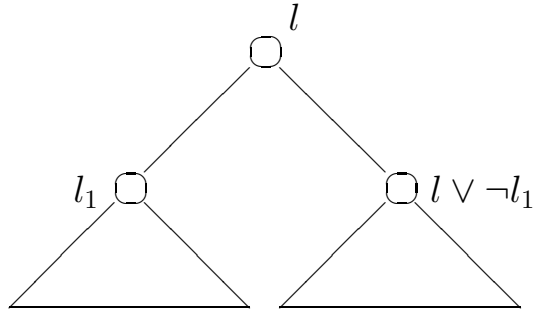

By induction, $\Pi \cup \{\neg l_1\} \models_{UP} \bot$, as the subtree rooted with $l_1$ have height less than or equal to $k$. By Lemma 1, $\neg l$ implies $\neg l_1$ by unit propagation. As a result, $\Pi \cup \{\neg l\} \models_{UP} \bot$.

Let us now consider the situation in which the children of $l$ are both binary clauses. Let us call $l_1$ the literal they are resolved upon, that is, the two clauses are $l \vee l_1$ and $l \vee \neg l_1$.

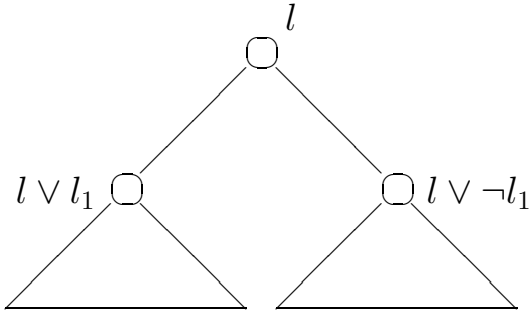

By Lemma 1, $\Pi \cup \{\neg l\} \models_{UP} l_1$ and $\Pi \cup \{\neg l\} \models_{UP} \neg l_1$. Since $\neg l$ allows deriving a pair of contradictory literals by unit propagation, we have $\Pi \cup \{\neg l\} \models_{UP} \bot$. □

We now use Property 3 to show how derivation is related to unit propagation.

**Lemma 3** *For every 2CNF formula $\Pi$ and literals $l_1$ and $l_2$, $\Pi \models l_1 \vee l_2$ if and only if $\Pi$ is inconsistent or $\Pi \cup \{\neg l_1\} \models_{UP} \bot$ or $\Pi \cup \{\neg l_2\} \models_{UP} \bot$ or $\Pi \cup \{\neg l_1\} \models_{UP} l_2$.*

*Proof.* The "if" direction is due to the fact the unit propagation is a sound (but not complete) entailment method, that is, if $\Pi \models_{UP} \gamma$ then $\Pi \models \gamma$.

The other direction is proved by applying Property 3. Indeed, $\Pi \models \gamma$ implies that either $\Pi$ is inconsistent, or it implies by resolution $l_1$, or $l_2$, or $l_1 \vee l_2$. In turns, $\Pi \models_R l_1$ implies $\Pi \cup \{\neg l_1\} \models_{UP} \bot$, and similarly for $\Pi \models_R l_2$ thanks to Lemma 2. Moreover, $\Pi \models_R l_1 \vee l_2$ implies $\Pi \cup \{\neg l_1\} \models_{UP} l_2$ thanks to Lemma 1. □

This lemma proves that unit propagation can be used to check whether a clause of two literals is implied by a consistent 2CNF formula. The following lemma is about inconsistent formulae.

**Lemma 4** *A 2CNF formula $\Pi$ is inconsistent if and only if there exists a variable $x$ such that $\Pi \cup \{x\} \models_{UP} \bot$ and $\Pi \cup \{\neg x\} \models_{UP} \bot$.*

*Proof.* The "if" direction is obvious, thanks to the soundness of unit propagation.

Let us consider a minimal regular resolution tree for $\Pi$. Its root is marked with $\bot$, so its children have to be marked $x$ and $\neg x$ for some variable $x$. Therefore, we have that $\Pi \models_R x$ and $\Pi \models_R \neg x$. By Lemma 2, the claim is proved. □

## 2.2 Formulae Implying Literals

A consistent 2CNF formula $\Pi$ can imply some literals or not. We show that, as long as redundancy and I.E.S.'s are concerned, we can threat the clauses containing an implied literal and those not containing them separately. We first formally prove that we can replace all clauses containing an entailed literal with the literal itself. This result is general to all formulae in CNF, and is a little obvious.

**Lemma 5** *If $\Pi$ is a CNF formula such that $\Pi \models l$, then $\Pi$ and $\Pi \backslash (\Pi|l) \cup \{l\}$ are equivalent.*

*Proof.* Let $M$ be a model of $\Pi$. By definition, $M$ satisfies all clauses of $\Pi$. Since $\Pi \models l$, this model assigns true to $l$. Since $\Pi \backslash (\Pi|l) \cup \{l\}$ only contains clauses of $\Pi$ and $l$, it is satisfied by $M$.

Vice versa, let $M$ be a model of $\Pi \backslash (\Pi|l) \cup \{l\}$. This model assigns true to $l$. Moreover, it satisfies all clauses of $\Pi$ not containing $l$. Since it also satisfies all clauses of $\Pi$ containing $l$ because it sets $l = \text{true}$, it satisfies all clauses of $\Pi$. □

Replacing all clauses containing an entailed literal with the literal itself does not only preserve equivalence but also redundancy, for 2CNF formulae.

**Lemma 6** *Let $\Pi$ be a formula implying the literal $l$. If the clause $\gamma \in \Pi$ does not contain $l$, then $\gamma$ is redundant in $\Pi$ if and only if it is redundant in $\Pi \backslash (\Pi|l) \cup \{l\}$.*

*Proof.* Since $\gamma$ does not contain $l$, it is both in $\Pi$ and in $\Pi \backslash (\Pi|l) \cup \{l\}$. We therefore only have to prove that is entailed by $\Pi \backslash \{\gamma\}$ if and only if it is entailed by $(\Pi \backslash (\Pi|l) \cup \{l\}) \backslash \{\gamma\}$. We prove that these two formulae are equivalent. Since $\gamma$ is not a clause of $\Pi|l$, we have that $(\Pi \backslash (\Pi|l) \cup \{l\}) \backslash \{\gamma\}$ is the same as $(\Pi \backslash \{\gamma\}) \backslash (\Pi|l) \cup \{l\}$. By Lemma 5, this formula is equivalent to $\Pi \backslash \{\gamma\}$. □

This lemma shows that clauses that do not contain literals that are entailed by the formula can be checked for redundancy after the clauses containing entailed literals have been removed. The following corollary is an obvious consequence of this lemma.

**Corollary 1** *If $\Pi$ is a consistent 2CNF such that $\Pi \models l$, then $\Pi$ contains a redundant clause not containing $l$ if and only if $\Pi \backslash (\Pi|l) \cup \{l\}$ is redundant.*

This lemma shows that, once we have determined that all clauses containing a literal $l$ such that $\Pi \models l$ are irredundant, we can replace all such clauses with $l$. Repeating this procedure for all literals implied by the formula, we obtain a formula with disjoint unit and binary clauses. This is because, if $\Pi \models l$, then every clause $l \vee l'$ is replaced by $l$, while every clause $\neg l \vee l'$ is replaced by $l'$. As a result, no binary clause contain a variable that is in a unit clause in the resulting formula.

We now show a similar result about I.E.S.'s, proving that clauses containing literals entailed by the formula can be replaced by these literals.

**Lemma 7** *Let $\Pi$ be a consistent 2CNF such that $\Pi \models l$. If $\Pi'$ is an I.E.S. of $\Pi$ then $\Pi_2 = \Pi' \backslash (\Pi|l) \cup \{l\}$ is an I.E.S. of $\Pi \backslash (\Pi|l) \cup \{l\}$.*

*Proof.* Since $\Pi'$ is an I.E.S. of $\Pi$, we have that $\Pi' \subseteq \Pi$. As a result, $\Pi' \backslash (\Pi|l) \cup \{l\} \subseteq \Pi \backslash (\Pi|l) \cup \{l\}$. Containment is the first condition for a formula being an I.E.S. of another formula. We now show equivalence and irredundancy.

Since $\Pi'$ is equivalent to $\Pi$, we have $\Pi' \models l$. By Lemma 5, $\Pi'$ is equivalent to $\Pi' \backslash (\Pi|l) \cup \{l\}$, which is indeed $\Pi_2$. As a result, $\Pi_2$ is equivalent to $\Pi'$, which is equivalent to $\Pi$, which is equivalent to $\Pi \backslash (\Pi|l) \cup \{l\}$ by Lemma 5.

Let us now assume that $\Pi' \backslash (\Pi|l) \cup \{l\}$ is redundant. The clause $l$ cannot be redundant because it is the only clause mentioning the literal $l$. Therefore, there exists a clause $\gamma$ not containing $l$ such that $\gamma$ is redundant in $\Pi' \backslash (\Pi|l) \cup \{l\}$. By Lemma 6, $\gamma$ is also redundant in $\Pi'$, thus contradicting the assumption that $\Pi'$ is an I.E.S. □

The converse of this lemma does not hold. Even if $\Pi_2$ is an I.E.S. of $\Pi \backslash (\Pi|l) \cup \{l\}$, it is not necessarily true that an I.E.S. of $\Pi$ can be obtained by simply adding some clauses of $\Pi|l$ to it. Actually, even adding adding *all* clauses of $\Pi|l$ does not necessarily lead to a formula that is equivalent to $\Pi$, as the following example shows.

$$\begin{aligned}
\Pi &= \{x \vee x_1, x \vee x_2, \neg x_1 \vee y, \neg x_2 \vee \neg y, \neg x \vee y\} \\
l &= x \\
\Pi \backslash (\Pi|l) \cup \{l\} &= \{\neg x_1 \vee y, \neg x_2 \vee \neg y, \neg x \vee y, x\} \\
\Pi_2 &= \{\neg x_2 \vee \neg y, \neg x \vee y, x\} \\
\Pi_2 \backslash \{l\} \cup (\Pi|l) &= \{\neg x_2 \vee \neg y, \neg x \vee y, x \vee x_1, x \vee x_2\}
\end{aligned}$$

It holds $\Pi \models x$: this can be proved by adding $\neg x$ to $\Pi$ and using unit propagation: $x_1$ and $x_2$ are derived, leading to $y$ and $\neg y$, respectively. It is also easy to prove that $\Pi_2$ is equivalent to $\Pi \backslash (\Pi|l) \cup \{l\}$: since $\neg x \vee y$ and $x$ entail $y$, the clause $\neg x_1 \vee y$ is entailed by $\Pi_2$. The irredundancy of $\Pi_2$ is also easy to prove: removing a single clause from it, either $x$, $y$, or $\neg x_2$ cannot be derived any longer. Adding all clauses of $\Pi|l$ to $\Pi_2$, however, does not allows to derive $x$ any longer. Indeed, $\Pi_2 \backslash \{l\} \cup (\Pi|l)$ has the model $\neg x x_1 x_2 \neg y$, which assigns false to $x$.

An analysis of this counterexample shows why the converse of Lemma 7 is false. The problem is with the clause $\neg x \vee y$, which makes $y$ true in the original formula. This clause allows to remove $\neg x_1 \vee y$, which was necessary in $\Pi$ to entail $x$. Without the clause $\neg x \vee y$, indeed, the converse of Lemma 7 would hold for $\Pi$ and $l = x$.

More precisely, a possible (but incorrect) proof of the converse of Lemma 7 would go by considering that $x_1$ and $x_2$ should always entail $y$ and $\neg y$, respectively, in every I.E.S. of $\Pi$, and therefore in any I.E.S. of $\Pi \backslash (\Pi|l) \cup \{l\}$, because this formula is equivalent to $\Pi$. What makes this proof fail on the counterexample above is that $x_1 \rightarrow y$ holds also because of $x$ and $\neg x \vee y$, and this proof relies on $x$, which is removed while "coming back" from $\Pi_2$ to $\Pi \backslash \{l\} \cup (\Pi|l)$.

As a result, the problem with this proof is that the clause $\neg x_1 \vee y$, which is part of the proof of $x$, is not necessary because it can be derived from $l = x$ in $\Pi \backslash (\Pi|l) \cup \{l\}$. On the other hand, $l$ can only derive clauses of the form $l \vee \textit{something}$, or $y \vee \textit{something}$ where $y$ is a consequence of $l$. In other words, what makes the proof fail is the possible entailment of clauses containing literals that are derivable from $l$. On the other hand, $\Pi \models l$; therefore, $\Pi \models y$. The counterexample therefore relies on the presence in $\Pi \backslash (\Pi|l) \cup \{l\}$ of clauses containing literals that are entailed by $\Pi$. Replacing all such clauses with that literal would therefore invalidate the counterexample. Since $\Pi_\models$ is the set of literals entailed by $\Pi$, the set $\Pi|\Pi_\models$ contains all clauses of $\Pi$ containing a literal that is entailed by $\Pi$.

**Lemma 8** *Every clause of a 2CNF formula $\Pi$ either is in $\Pi|\Pi_\models$ or does not contain literals in $\Pi_\models$ or their negation.*

*Proof.* Let $l \in \Pi_\models$. All clauses containing $l$ are in $\Pi|\Pi_\models$ by definition. On the other hand, all clauses containing the negation of $l$ are in the form $\neg l \vee l'$. Since $\Pi \models l$, we have $\Pi \models l'$. Therefore $l \vee l' \in \Pi|\Pi_\models$. □

The following is an obvious consequence of the above lemma.

**Corollary 2** *For every 2CNF formula $\Pi$, it holds that $\Pi \backslash (\Pi|\Pi_\models)$ and $\Pi_\models$ do not share variables.*

This corollary is the base of the next result. Indeed, it shows that $\Pi \backslash (\Pi|\Pi_\models) \cup \Pi_\models$ is composed of two completely separated parts $\Pi \backslash (\Pi|\Pi_\models)$ and $\Pi_\models$. The same therefore holds for any of its subsets, and in particular for all of its I.E.S.'s. We show the following lemma proving that replacing *at once* all clauses of $\Pi|\Pi_\models$ with $\Pi_\models$, the converse of Lemma 7 holds.

**Lemma 9** *Let $\Pi$ be a consistent 2CNF formula. If $\Pi_2$ is an I.E.S. of $\Pi \backslash (\Pi|\Pi_\models) \cup \Pi_\models$, then $\Pi_2 \backslash \Pi_\models \cup (\Pi|\Pi_\models)$ is equivalent to $\Pi$.*

*Proof.* Since $\Pi$ and $\Pi\backslash(\Pi|\Pi_\models) \cup \Pi_\models$ are equivalent and $\Pi_2$ is an I.E.S. of $\Pi\backslash(\Pi|\Pi_\models) \cup \Pi_\models$, we have that $\Pi_2$ and $\Pi$ are equivalent. The claim is proved by showing that $\Pi_2\backslash\Pi_\models \cup (\Pi|\Pi_\models)$ entails $\Pi_\models$. This would prove that $\Pi_2\backslash\Pi_\models \cup (\Pi|\Pi_\models)$ is equivalent to $\Pi_2\backslash\Pi_\models \cup (\Pi|\Pi_\models) \cup \Pi_\models$, which is a superset of $\Pi_2$ and is therefore equivalent to $\Pi$.

Intuitively, the proof is as follows: if $l \in \Pi_\models$, then there is a proof of $l$ in $\Pi$. This proof involves some clauses of $\Pi|\Pi_\models$ and some clauses in $\Pi\backslash(\Pi|\Pi_\models)$. On the other hand, everything that is entailed in $\Pi$ is also entailed in its equivalent formula $\Pi_2\backslash(\Pi|\Pi_\models) \cup \Pi_\models$. Since $\Pi_2\backslash(\Pi|\Pi_\models)$ and $\Pi_\models$ are built over disjoint literals, every clause of $\Pi$ not containing literals that are entailed by $\Pi$ is derivable in $\Pi_2\backslash(\Pi|\Pi_\models)$.

Let us formally prove the claim. Let $l \in \Pi_\models$, that is, $\Pi \models l$. We show that $\Pi_2\backslash\Pi_\models \cup (\Pi|\Pi_\models) \models l$. Since $\Pi \models l$, two conditions are possible: either $l \in \Pi$, or $l \notin \Pi$. In the first case, $l \in \Pi|\Pi_\models$, and the claim is true because $l$ is in $\Pi_2\backslash\Pi_\models \cup (\Pi|\Pi_\models)$.

Let us now consider the case $l \notin \Pi$. Since $\Pi \models l$ and $\Pi$ is consistent, $\Pi \cup \{\neg l\}$ allows deriving a pair of opposite literals by unit propagation. Let the following be the chains of clauses used in the unit propagation from $\neg l$ to these pairs of opposite literals:

$$p_1 \to p_2 \to \cdots \to p_{n-1} \to p_n$$
$$n_1 \to n_2 \to \cdots \to n_{n-1} \to \neg n_n$$
$$\text{where } \neg l = p_1 = n_1 \text{ and } n_n = \neg p_n$$

Consider an arbitrary link $l_i \to l_{i+1}$ of these two chains, corresponding to the clause $\neg l_i \vee l_{i+1}$. By Lemma 8, either this clause is in $\Pi|\Pi_\models$ or it does not share variables with $\Pi_\models$. Let us now consider this second case.

Since this clause is in $\Pi$, it holds $\Pi_2 \models l_i \to l_{i+1}$. On the other hand, $\Pi_2 = (\Pi_2\backslash\Pi_\models) \cup (\Pi_2 \cap \Pi_\models)$. These two parts of $\Pi_2$ contains disjoint literals because of Corollary 2. Since $\Pi_2 \models l_i \to l_{i+1}$ and neither $l_i$, $l_{i+1}$, nor their negations are in $\Pi_\models$, then $\Pi_2\backslash\Pi_\models \models l_i \to l_{i+1}$ because the other part of $\Pi_2$ contains literals that are mentioned neither in $\Pi_2\backslash\Pi_\models$ nor in $l_i \to l_{i+1}$.

Of the clauses of the two chains above, therefore, we have that a clause $l_i \to l_{i+1}$ is either in $\Pi|\Pi_\models$ or is entailed by $\Pi_2\backslash\Pi_\models$. As a result, $\Pi_2\backslash\Pi_\models \cup (\Pi|\Pi_\models)$ entails all these clauses, and therefore the unit propagation from $\neg l$ leads to two pair of opposite literals in this formula. □

This lemma only proves that $\Pi_2\backslash\Pi_\models \cup (\Pi|\Pi_\models)$ is equivalent to $\Pi$, but does not prove it is an I.E.S. of $\Pi$. In general, this is not true. However, we can show that an I.E.S. can be obtained from this set by removing only some clauses of $\Pi|\Pi_\models$.

**Lemma 10** *Let $\Pi$ be a consistent 2CNF formula. If $\Pi_2$ is an I.E.S. of $\Pi\backslash(\Pi|\Pi_\models) \cup \Pi_\models$, then there exists $\Pi_1 \subseteq \Pi|\Pi_\models$ such that $\Pi_1 \cup (\Pi_2\backslash\Pi_\models)$ is an I.E.S. of $\Pi$.*

*Proof.* Lemma 9 shows that adding $\Pi|\Pi_\models$ to $\Pi_2\backslash\Pi_\models$ results in a set that is equivalent to $\Pi$. We are now trying to prove that an I.E.S. of $\Pi$ can be obtained from $\Pi_2\backslash\Pi_\models \cup (\Pi|\Pi_\models)$ without removing any clause of $\Pi_2\backslash\Pi_\models$. What we actually prove is that no clause of $\Pi_2\backslash\Pi_\models$ is redundant in $\Pi_2\backslash\Pi_\models \cup (\Pi|\Pi_\models)$, thus proving that all I.E.S.'s of $\Pi_2\backslash\Pi_\models \cup (\Pi|\Pi_\models)$ contain all clauses of $\Pi_2\backslash\Pi_\models$ by Lemma 1.

Let $\gamma \in \Pi_2 \backslash \Pi_\models$. Assume that $\gamma$ is redundant in $\Pi_2 \backslash \Pi_\models \cup (\Pi | \Pi_\models)$, that is:

$$(\Pi_2 \backslash \Pi_\models \cup (\Pi | \Pi_\models)) \backslash \{\gamma\} \models \gamma$$

Since all clauses in $\Pi | \Pi_\models$ contains a literal in $\Pi_\models$ by definition, we have that $\Pi_\models \models \Pi | \Pi_\models$. As a result, $\Pi_\models$ is logically stronger than $\Pi | \Pi_\models$, and the above formula therefore implies:

$$(\Pi_2 \backslash \Pi_\models \cup \Pi_\models) \backslash \{\gamma\} \models \gamma$$

This is the same as $\Pi_2 \backslash \{\gamma\} \models \gamma$, contradicting the assumption that $\Pi_2$ is irredundant. $\square$

The converse of this lemma is an immediate consequence of a repeated application of Lemma 7. We can therefore conclude the following corollary.

**Corollary 3** $\Pi_2$ *is an* I.E.S. *of* $\Pi \backslash (\Pi | \Pi_\models) \cup \Pi_\models$ *if and only if there exists* $\Pi_1 \subseteq \Pi | \Pi_\models$ *such that* $\Pi_1 \cup (\Pi_2 \backslash \Pi_\models)$ *is an* I.E.S. *of* $\Pi$.

## 2.3 Cyclicity and Induced Graphs

The presence of cycles of clauses in a formula determines the complexity of some problems related to I.E.S.'s. Formally, cycles are defined as follows.

**Definition 4 (Simple Cycle of Binary Clauses)** *A cycle of binary clauses is a sequence of clauses* $[\neg l_1 \vee l_2, \neg l_2 \vee l_3, \ldots, \neg l_n \vee l_1]$ *such that no literal* $l_i$ *occur in more than two clauses.*

This definition only covers simple cycles of clauses, that is, we are not allowed to "cross" the same literal twice. A non-simple cycle of clauses can be defined as a sequence $[\neg l_1 \vee l_2, \neg l_2 \vee l_3, \ldots, \neg l_n \vee l_1]$ in which there is no pair of indexes $i, j$ with $i \neq j$ such that $l_j = \neg l_i$. This definition is however not necessary because we only classify formulae based on whether they have cycles or not. Since every formula having cycles also have simple cycles (and, obviously, the other way around), the classification based on having or not having simple cycles is sufficient.

The graph of a 2CNF formula induced by a literal is, roughly speaking, the graph of literals that can be derived from the given one by unit propagation.

**Definition 5** *The graph induced by a literal $l$ on a 2CNF formula $\Pi$ is the minimal graph such that:*

1. *$l$ is a node of the graph;*

2. *if $l'$ is a node of the graph and $\neg l' \vee l'' \in \Pi$, then $l''$ is a node of the graph and $(l', l'')$ is an edge of the graph.*

A property of acyclic formulae is that all its induced subgraphs are acyclic and vice versa.

**Property 4** *A 2CNF formula $\Pi$ contains simple cycles if and only if some of its induced graphs contain cycles.*

Formulae not containing cycles have some interesting properties. First, consistent acyclic 2CNF formulae not entailing any literal have a single I.E.S. This result is interesting also because it makes some problems related to I.E.S.'s computationally simpler. The second result about acyclic formulae is that, as far as two literals $l_1$ and $l_2$ that are entailed by the formula are concerned, the choice of a minimal subset of clauses entailing $l_1$ can be done independently of the choice for $l_2$. These two results are proved in two later sections.

## 2.4 Unit Clauses

If $\Pi$ only contains binary clauses, then $\Pi \cup \{l\} \models_{UP} l'$ has the only one possible meaning that $l'$ is obtained from $\Pi$ by applying unit propagation starting from $l$ because $l$ is the only unit clause of $\Pi \cup \{l\}$. If this is the case, $\Pi \cup \{l\} \models_{UP} l'$ is equivalent to the reachability of $l'$ from $l$ in the graph of $\Pi$ induced by $l$.

In most cases, however, $\Pi$ cannot be assumed to be composed of binary clauses only. In particular, even if we start from a formula made of binary clauses only, applying Corollary 1 or Lemma 3 leads to formulae containing unit clauses. On the other hand, every unit clause $l$ can be replaced with the logically equivalent pair $\{l \vee l', l \vee \neg l'\}$, where $l'$ is a new variable not occurring in the rest of the formula. Since $\{l\} \equiv \{l \vee l', l \vee \neg l'\}$, the redundancy of $l$ is equivalent to the redundancy of the pair $\{l \vee l', l \vee \neg l'\}$, and most of the properties related to I.E.S. are also unaffected by this replacement.

The only property that is changed by replacing $l$ with $\{l \vee l', l \vee \neg l'\}$ is about the size of I.E.S.'s of a formula, as we are replacing a single clause with a pair of clauses. This problem will be taken care by counting such a pair as if it were a single clause in the algorithms for checking the size of a minimal I.E.S. of a formula.

In the rest of this paper, whenever we have to check whether $\Pi \cup \{l_1\} \models_{UP} l_2$ or $\Pi \cup \{l_1\} \models_{UP} \bot$, we assume that this transformation has been preliminary been done on $\Pi$, so that $\Pi$ only contains binary clauses. This way, checking unit propagation can be done by looking at the graph of $\Pi$ induced by $l_1$. In particular, $\Pi \cup \{l_1\} \models_{UP} l_2$ means that, in the graph of $\Pi$ induced by $l_1$, there is a path from $l_1$ to $l_2$.

By definition, $\Pi \cup \{l_1\} \models_{UP} \bot$ means that unit propagation from $l$ allows reaching a pair of opposite literals $l_2$ and $\neg l_2$. Graphically, there exists a path from $l_1$ to $l_2$ and a path from $l_1$ to $\neg l_2$ in the graph of $\Pi$ induced by $l_1$. Let $l_3$ be the last common literal of these two paths.

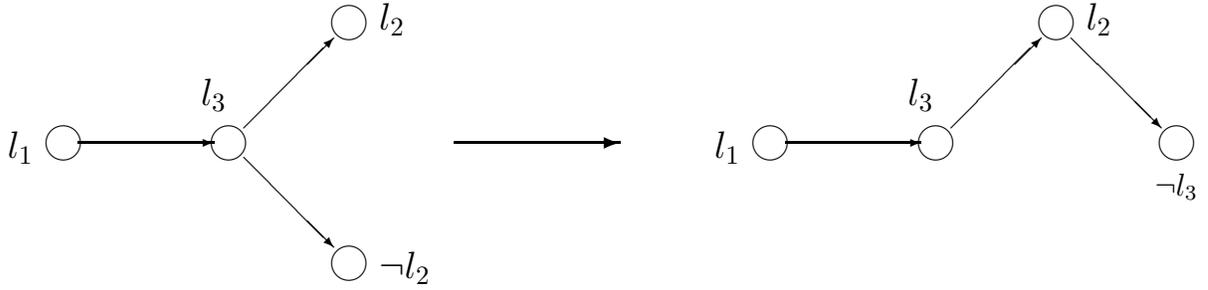

Since $\neg l_2$ can be reached from $l_3$, we have that $\neg l_3$ can be reached from $l_2$. As a result, there exists a path from $l_1$ to $l_3$, from $l_3$ to $l_2$, and from $l_2$ to $\neg l_3$. In other words, $\Pi \cup \{l_1\} \models_{UP} \perp$ implies that there is a *single* path starting from $l_1$ and that includes a pair of opposite literals. In the sequel, whenever we say "all clauses in a path from $l_1$ to $\perp$", we mean a single path starting from $l_1$ and ending with *the first* literal that is the opposite of another literal in the path.

## 2.5 The Three Cases

As reported in the Introduction, three cases are studied separately, both for the problem of redundancy checking and the problems about I.E.S.'s:

1. the formula is inconsistent;

2. the formula is consistent and implies some literals;

3. the formula is consistent and does not imply literals.

We can now explain why these three cases require a different analysis. Let us consider first the last case: a formula not implying any literal. By Lemma 3, entailment of a clause hold if and only if one of its literals can be derived by unit propagation from the negation of the other one. The same must therefore be true for all I.E.S.'s of the formula. As a result, the problem of redundancy and the problems about I.E.S.'s can be reformulated in terms of formulae whose induced graphs have the same reachability relation of the graphs of the original formula.

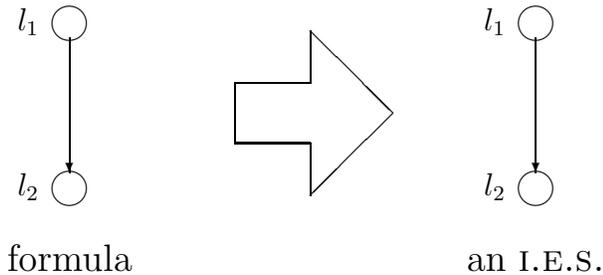

formula          an I.E.S.

This requirement, however, is too restrictive when considering literals that are implied by the formula. For example, if $l_2$ is reachable from $l_1$ in the original formula but $\neg l_1$ is entailed by the formula, this reachability condition is not necessarily true in all I.E.S.'s of the formula. Indeed, all that is needed is that unit propagation from $l_1$ allows reaching contradiction; the condition that $l_2$ is reachable from $l_1$ is not necessarily true in all I.E.S.'s of the formula.

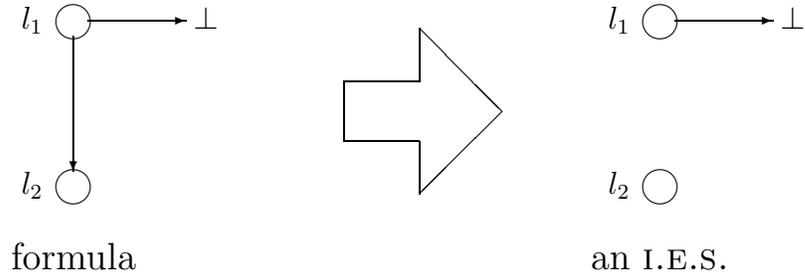

formula                          an I.E.S.

Finally, if a formula is inconsistent then its I.E.S.'s are only required to be inconsistent. Not only these I.E.S.'s are no longer required to have the same reachability relation of the original formula: they can even omit to mention some literals at all. Indeed, a subset of an inconsistent formula can be inconsistent even if it does not mention some literals of the original formula.

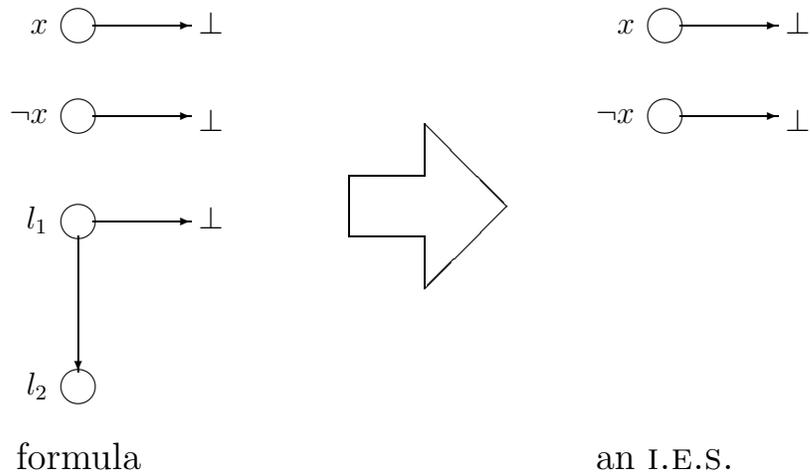

formula                          an I.E.S.

Summarizing, the three cases are studied separately because of the different requirement on equivalent subsets: in the first case, reachability using unit propagation is the same in the I.E.S.'s and in the formula; in the second case, only reachability of $\bot$ from the negation of implied literals is the same; in the third case, only the reachability of $\bot$ from an arbitrary pair of opposite literals is the same.

## 2.6 Number of Clauses

We show a lemma about clauses containing a literal that is entailed by the formula.

**Lemma 11** *Every consistent 2CNF formula $\Pi$ such that $\Pi \models l$ and containing three or more clauses containing $l$ has at least a redundant clause containing $l$.*

*Proof.* If $l \in \Pi$, any other clause containing $l$ is redundant. Let us assume $l \notin \Pi$. Since $\Pi$ is consistent, $\Pi \models l$ is equivalent to $\Pi \cup \{\neg l\} \models_{UP} \bot$. By definition, this formula is true if and only if $\neg l$ allows deriving a pair of complementary literals $x$ and $\neg x$ by unit propagation.

Let us first assume that neither $l$ is neither $x$ nor $\neg x$. Consider the sequences of clauses used in the derivation of $x$ and $\neg x$ from $l$. Without loss of generality, we can assume that $l$ is only contained in the first clauses of these two sequences.

Removing all clauses containing $l$ but the first clause in each of these two sequences, we obtain a formula that still entails $l$, and therefore allows deriving all clauses that have been removed, contradicting the assumption that $\Pi$ is irredundant.

A similar proof can be used for the case in which $l = x$ or $l = \neg x$. In this case, the first clause of the sequence of clauses used in the derivation of $\neg l$ from $l$ allows deriving all other clauses containing $l$. □

An obvious consequence of this lemma is that every consistent and irredundant 2CNF formula implying a literal contains at most two clauses containing that literal.

## 2.7 Acyclic Consistent 2CNF Formulae not Implying Literals Have a Single I.E.S.

We show that every acyclic consistent 2CNF formula not implying single literals has a single I.E.S. In this section, we assume that $\Pi$ is a 2CNF formula that is consistent, acyclic, and it does not imply any single literal. Since $\Pi$ is consistent and $\Pi \not\models l$ for every literal $l$, we have that $\Pi \models \neg l \vee l'$ holds if and only if $\Pi \cup \{l\} \models_{UP} l'$. The same holds for all its subsets and, in particular, for all its I.E.S.'s.

**Lemma 12** *If $\Pi \cup \{l\} \models_{UP} l'$ then $\Pi' \cup \{l\} \models_{UP} l'$ holds for every I.E.S. $\Pi'$ of the consistent acyclic 2CNF formula $\Pi$.*

*Proof.* Since $\Pi \cup \{l\} \models_{UP} l'$ it holds $\Pi \models \neg l \vee l'$, and the same therefore holds for $\Pi'$ because this formula is equivalent to $\Pi$. On the other hand, $\Pi'$ is consistent and does not entail literals because so is $\Pi$. As a result, $\Pi' \models \neg l \vee l'$ is equivalent to $\Pi' \cup \{l\} \models_{UP} l'$. □

Every literal $l$ partitions $\Pi$ into the set of clauses that are involved in the first step of unit propagation from $l$ and the other ones:

$$\begin{aligned} D_\Pi(l) &= \{\gamma \in \Pi \mid \gamma = \neg l \vee l'\} \\ R_\Pi(l) &= \Pi \backslash D_\Pi(l) \end{aligned}$$

The clauses in $D_\Pi(l)$ are those used in the first step of unit propagation from $l$. The following set $C_\Pi(l)$ is the set of literals that would result from this propagation.

$$C_\Pi(l) = \{l' \mid \neg l \vee l' \in \Pi\} = \{l' \mid \neg l \vee l' \in D_\Pi(l)\}$$

Since all clauses $\neg l \vee l'$ in $D_\Pi(l)$ are also in $\Pi$, they are entailed by every I.E.S. $\Pi'$ of $\Pi$. Since $\Pi$ is a consistent CNF, so are all its subsets, and $\Pi'$ in particular. Since $\Pi'$ entails $\neg l \vee l'$ and $\Pi'$ is consistent and not entailing literals, $l'$ is reachable from $l$ in the graph induced by $l$ on $\Pi'$.

In turn, a given $l' \in C_\Pi(l)$ is reachable from $l$ if and only if either $\neg l \vee l' \in \Pi'$, or there is another literal $l''$ such that $\neg l \vee l'' \in \Pi'$ and $l'$ is reachable from $l''$ using the edges corresponding to the clauses of $R_\Pi(l)$. Consider the following graph induced by $l$ on a 2CNF formula.

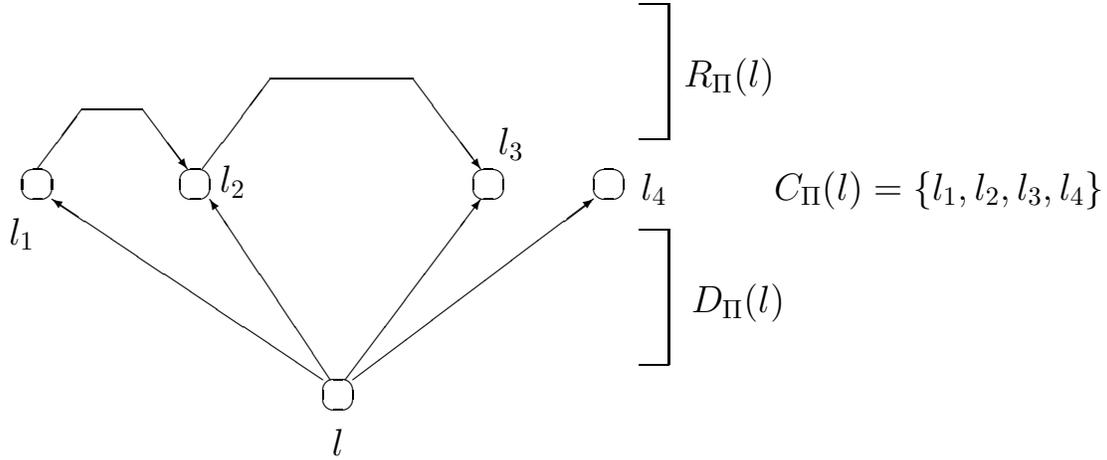

In this example, $\neg l \vee l_4$ cannot be removed from $\Pi$ because it is the only clause allowing $l_4$ to be reached from $l$. The same holds for $\neg l \vee l_1$. The clauses $\neg l \vee l_2$ and $\neg l \vee l_3$ are instead redundant because $l_2$ and $l_3$ can be reached from $l$ by following the edge corresponding to the clause $\neg l \vee l_1$ and then following some edges corresponding to clauses in $R_\Pi(l)$.

This example shows that the irredundant clauses are those containing the literals of $C_\Pi(l)$ that cannot be reached from other literals of $C_\Pi(l)$. Such literals necessarily exist because the formula (and therefore all its induced graphs) are acyclic.

$$M_\Pi(l) = \{l' \in C_\Pi(l) \mid \nexists l'' \in C_\Pi(l) \text{ such that } R_\Pi(l) \cup \{l''\} \models_{UP} l'\}$$

Formally, $M_\Pi(l)$ is the set of literals that cannot be reached from other literals of $C_\Pi(l)$ using unit propagation on the clauses $R_\Pi(l)$. In the example above, $R_\Pi(l)$ and $M_\Pi(l)$ are as follows:

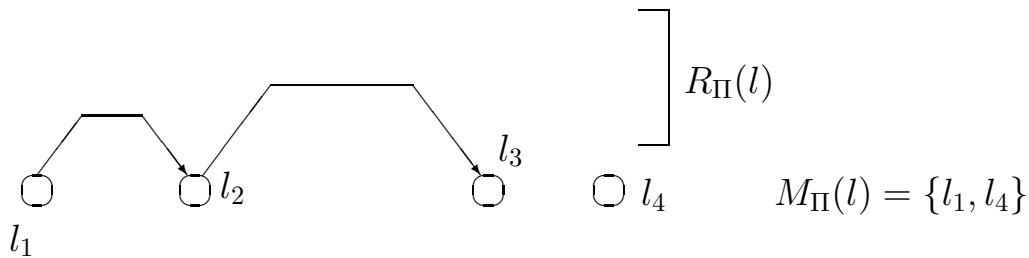

The nodes that cannot be reached from from other nodes are $l_1$ and $l_4$. Therefore, $M_\Pi(l) = \{l_1, l_4\}$. The next two lemmas formally prove that $M_\Pi(l)$ exactly characterizes the clauses containing $\neg l$ that are in some I.E.S.'s.

**Lemma 13** *If $\Pi$ is a consistent acyclic 2CNF formula not implying literals and $l_1 \in M_\Pi(l)$ then any I.E.S. of $\Pi$ contains $\neg l \vee l_1$.*

*Proof.* Let us assume that $\Pi'$ is an I.E.S. of $\Pi$. Since $\Pi \cup \{l\} \models_{UP} l_1$, we have that $\Pi' \cup \{l\} \models_{UP} l_1$ by Lemma 12. Let us assume that $\neg l \vee l_1$ is not in $\Pi'$. Then, the path from $l$ to $l_1$ is made of an edge corresponding to a clause $\neg l \vee l_2$ with $l_2 \neq l_1$ followed by a path from $l_2$ to $l_1$. This path cannot include $l$ because otherwise $\Pi$ would be cyclic. Therefore, this path from $l_2$ to $l_1$ is all contained in $R_\Pi(l)$. This is however in contradiction with $l_1$ being in $M_\Pi(l)$. □

The second lemma is the converse of the previous one, stating that literals not in $C_\Pi(l)$ do not form clauses with $l$ that are in any I.E.S.

**Lemma 14** *If $l_1 \in C_\Pi(l) \backslash M_\Pi(l)$ then no I.E.S. of the consistent acyclic 2CNF formula $\Pi$ contains $\neg l \vee l_1$.*

*Proof.* By definition of $M_\Pi(l)$, if a literal $l_1$ is in $C_\Pi(l)$ but not in $M_\Pi(l)$, then there is a literal $l'$ in $M_\Pi(l)$ such that $R_\Pi(l)$ contains a path from $l'$ to $l_1$. Let $\Pi'$ be a I.E.S. of $\Pi$. Since $l_1$ is reachable from $l'$, it holds $\Pi \cup \{\neg l'\} \models_{UP} l_1$ and therefore $\Pi' \cup \{\neg l'\} \models_{UP} l_1$.

By the previous lemma, $\Pi'$ contains $\neg l \vee l'$. As a result, $\Pi' \cup \{l\} \models_{UP} l'$. Since $\Pi' \cup \{\neg l'\} \models_{UP} l_1$, we have that $l_1$ can be reached from $l$ by first using the clause $\neg l \vee l'$. Since the formula is acyclic, the clauses used in the derivation $\Pi' \cup \{\neg l'\} \models_{UP} l_1$ cannot contain $l$. As a result, unit propagation allows to reach $l_1$ from $l$ without using the clause $\neg l \vee l_1$. In other words, $\Pi' \backslash \{\neg l \vee l_1\} \cup \{l\} \models_{UP} l_1$, thus proving that $\neg l \vee l_1$ is redundant in $\Pi'$, contradicting the assumption that $\Pi'$ is an I.E.S. □

We can then conclude that $M_\Pi(l)$ exactly identifies all clauses of $D_\Pi(l)$ that are in an I.E.S. of $\Pi$.

**Corollary 4** *A clause $\neg l \vee l_1$ is in an I.E.S. of the consistent acyclic 2CNF formula $\Pi$ if and only if $l_1 \in M_\Pi(l)$.*

Once $l_1 \vee l_2$ is proved to be in an I.E.S. because $l_2 \in M_\Pi(\neg l_1)$, we do not need to also check $l_1 \in M_\Pi(\neg l_2)$. The same holds if $l_1 \vee l_2$ is proved not to be in an I.E.S. in the same way. This result tells that every 2CNF consistent acyclic formula that does not imply literals has a single I.E.S.

**Theorem 1** *Every consistent acyclic 2CNF formula $\Pi$ not implying literals has a single I.E.S.*

*Proof.* For each clause $l_1 \vee l_2$, check whether $l_2 \in M_\Pi(\neg l_1)$. If this is true, then $l_1 \vee l_2$ is in all I.E.S.'s, otherwise it is in no I.E.S. Therefore, the set composed of all clauses $l_1 \vee l_2$ such that $l_2 \in M_\Pi(\neg l_1)$ is the single I.E.S. of $\Pi$. □

## 2.8 Implied Literals not in a Cycle of a Consistent 2CNF Formula

We consider the clauses containing literals that are entailed by the formula but are not in a cycle of clauses of the formula. We show that, if $l'$ and $l''$ are two such literals, then we can independently choose among clauses containing $l'$ and from clauses containing $l''$ to form an I.E.S. In other words, an I.E.S. can be obtained by choosing a subset of clauses containing $l'$ and a subset of clauses containing $l''$, and these choices are independent from each other.

By Lemma 2, since $\Pi$ is consistent, $\Pi \models \neg l$ holds if and only if $\Pi \cup \{l\} \models_{UP} \bot$. Using the transformation of Section 2.4, we can assume that $\Pi$ does not contain unary clauses, and therefore $\Pi \cup \{l\} \models_{UP} \bot$ means that we can reach a pair of opposite literals by propagating $l$ in $\Pi$. We partition the clauses of $\Pi$ in those containing $\neg l$ and those which does not, and define the set of literals that are direct consequences of $l$.

$$\begin{aligned}
D_\Pi(l) &= \{\gamma \in \Pi \mid \gamma = \neg l \vee l'\} \\
R_\Pi(l) &= \Pi \backslash D_\Pi(l) \\
C_\Pi(l) &= \{l' \mid \neg l \vee l' \in \Pi\}
\end{aligned}$$

Since $\Pi \models \neg l$, every equivalent subset of $\Pi$ allows reaching a pair of opposite literals from $l$. This is possible if and only if either one of the two following conditions is true:

1. $\neg l \vee l_1 \in \Pi$ and $\Pi \cup \{l_1\} \models_{UP} \bot$;

2. $\neg l \vee l_2 \in \Pi$, $\Pi \cup \{l_2\} \models_{UP} \neg l$, which is equivalent to:
   $\neg l \vee l_2 \in \Pi$, $\Pi \cup \{l_2\} \models_{UP} \neg l_3$, and $l_3 \vee \neg l \in \Pi$.

If $\Pi'$ is an I.E.S. of $\Pi$, these conditions hold for $\Pi$ if and if they hold for $\Pi'$. In addition, if $\Pi$ does not contain a cycle including $l$, the same hold for $\Pi'$. As a result, the unit propagation from $l_1$ to $\bot$ or from $l_2$ to $\neg l_3$ cannot include $l$, as otherwise $l$ would be part of a cycle. We show a skecth of the proof on the formula represented by the following figure.

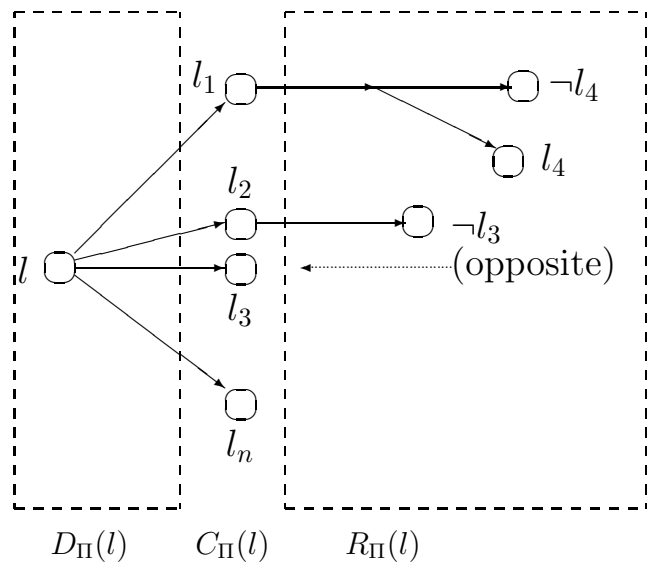

All paths from $l_1$ to $\bot$ and from $l_2$ to $\neg l_3$ are entirely contained in $R_\Pi(l)$: otherwise, $l$ would be part of a cycle. As a result, the same derivations are possible in an I.E.S. $\Pi'$ of $\Pi$ if and only if they are possible using only clauses of $R_\Pi(l)$, that is, they are possible in $\Pi' \cap R_\Pi(l)$. In other words, $\Pi' \cup \{l_1\} \models_{UP} \bot$ if and only if $(\Pi' \cap R_\Pi(l)) \cup \{l_2\} \models_{UP} \bot$, and the same for the derivation of $\neg l_3$ from $l_2$. As a result, $\Pi' \cap R_\Pi(l)$ always entails $l$ with the addition of either $\neg l \vee l_1$ or the pair $\neg l \vee l_2$ and $\neg l \vee l_3$.

Let us formally define the literals like $l_1$ and the pairs of literals like $l_2$ and $l_3$.

$$S_\Pi(l) = \{l_1 \in C_\Pi(l) \mid R_\Pi(l) \cup \{l_1\} \models_{UP} \bot\}$$
$$P_\Pi(l) = \{(l_2, l_3) \in C_\Pi(l) \mid R_\Pi(l) \cup \{l_2\} \models_{UP} \neg l_3\} \backslash S_\Pi(l)$$

In words, $S_\Pi(l)$ is the set of literals we can reach contradiction from using unit propagation in $R_\Pi(l)$ while $P_\Pi(l)$ is composed of the pair of literals such that the negation of one is reachable from the other in $R_\Pi(l)$.

**Lemma 15** *If $\Pi'$ is an I.E.S. of the consistent acyclic 2CNF formula $\Pi$ such that $\Pi \models \neg l$ and $l$ is not in any cycle of clauses of $\Pi$, then the following are all I.E.S.'s of $\Pi$:*

1. $(\Pi' \cap R_\Pi(l)) \cup \{\neg l \vee l_1\}$ *with* $l_1 \in S_\Pi(l)$;

2. $(\Pi' \cap R_\Pi(l)) \cup \{\neg l \vee l_2, \neg l \vee l_3\}$ *with* $(l_2, l_3) \in P_\Pi(l)$.

*Proof.* $\Pi' \cap R_\Pi(l)$ contains all clauses of $\Pi'$ but those containing $\neg l$. As a result, if we can prove that the formulae above entail $\neg l$, that would prove that they are equivalent to $\Pi'$.

Let us consider $(\Pi' \cap R_\Pi(l)) \cup \{\neg l \vee l_1\}$ first. Since $l_1 \in S_\Pi(l)$, we have that $\Pi \cup \{l_1\} \models_{UP} \bot$. Since $l$ is not part of a cycle, all derivations of $\bot$ from $l_1$ are enterely contained in $R_\Pi(l)$. Since $\Pi'$ is equivalent to $\Pi$, it holds $\Pi' \cup \{l_1\} \models \bot$; since $\Pi'$ is a subset of $\Pi$, all derivations of $\bot$ from $l_1$ only use clauses of $R_\Pi(l)$. As a result, $(\Pi' \cap R_\Pi(l)) \cup \{l_1\} \models \bot$, which implies that $(\Pi' \cap R_\Pi(l)) \cup \{\neg l \vee l_1\} \cup \{l\} \models \bot$.

Since $\Pi'$ is irredundant, $\Pi' \cap R_\Pi(l)$ is irredundant as well. In order to prove that $(\Pi' \cap R_\Pi(l)) \cup \{\neg l \vee l_1\}$ is irredundant, observe that $\neg l \vee l_1$ is not redundant because $\Pi' \cap R_\Pi(l)$ do not contain clauses containing $\neg l$ and therefore cannot entail $\neg l$. Regarding the clauses of $\Pi' \cap R_\Pi(l)$, since they do not contain $\neg l$ by definition, Lemma 6 applies: they are redundant in $(\Pi' \cap R_\Pi(l)) \cup \{\neg l \vee l_1\}$ if and only if they are redundant in $\Pi' \cap R_\Pi(l)$, which is impossible because $\Pi'$ is irredundant.

The proof for $(\Pi' \cap R_\Pi(l)) \cup \{\neg l \vee l_2, \neg l \vee l_3\}$ with $(l_2, l_3) \in S_\Pi(l)$ is similar. This formula implies $\neg l$ because $\neg l_3$ is reachable from $l_2$, and all paths from $l_2$ to $\neg l_3$ are in $R_\Pi(l)$. Since $\Pi'$ is equivalent to $\Pi$, it contains one such path, that is therefore all contained in $\Pi' \cap R_\Pi(l)$. As a result, the addition of the clauses $\neg l \vee l_2$ and $\neg l \vee l_3$ allows the entailment of $l$.

The proof of irredundancy of $(\Pi' \cap R_\Pi(l)) \cup \{\neg l \vee l_2, \neg l \vee l_3\}$ is also similar to the previous one. However, for this proof to work we also need the fact that neither $l_2$ nor $l_3$ are in $S_\Pi(l)$, and therefore one of them is not sufficient for entailing $l$. □

This lemma shows a simple way for determining an I.E.S. of an acyclic consistent formula: for each literal $l$ such that $\Pi \models \neg l$, we choose either a clause $l \vee l_1$ with $l_1 \in S_\Pi(l)$ or a pair of clauses $l \vee l_2$ and $l \vee l_3$ with $(l_2, l_3) \in P_\Pi(l)$. By the above lemma, we can do this choice for all clauses containing a literal that is entailed by the formula.

# 3  Redundancy Checking

In this section, we consider the problem of checking whether a 2CNF formula is redundant. The trivial algorithm of checking whether $\Pi\backslash\{\gamma\} \models \gamma$ for each $\gamma \in \Pi$ takes time $O(mm)$, where $m$ is the number of clauses in $\Pi$. We improve over this result by showing algorithms that solve the problem in time $O(nm)$, where $n$ is the number of variables, in all three possible cases (inconsistent formulae, formulae implying literals or not.) Cyclicity does not affect the problem of checking redundancy. Here is a summary of the results in the various cases.

**The formula is inconsistent.** If a set is both inconsistent and irredundant, the number of its clauses is at most four times the number of variables. Therefore, if the number of clauses is greater, the set is redundant. If it is lower, we still have to check redundancy, but the running time $O(mm)$ of the trivial algorithm is now the same as $O(nm)$.

**The formula is consistent.** For each literal $l$, we proceed differently depending on whether $\Pi \models l$ or not.

> **The formula implies the literal.** If $\Pi \models l$ then either $\Pi$ is inconsistent or $\Pi \models_R l$. Since $\Pi$ is by assumption consistent, we have $\Pi \models_R l$. By Lemma 2, we have $\Pi \cup \{\neg l\} \models_{UP} \bot$. We consider two cases separately.
>
>> $l$ **is in three or more clauses.** Formula $\Pi$ is redundant by Lemma 11.
>>
>> $l$ **is in one or two clauses.** By assumption $\Pi \cup \{\neg l\} \models_{UP} \bot$. In order to check redundancy, just remove any of the two clauses containing $l$ and check whether UP still leads to $\bot$ from $l$. Two UP derivations, which are linear in the number of clauses, are needed for the literal $l$.
>
> **The formula does not imply the literal.** Since $\Pi \not\models l$ we have $\Pi\backslash\{\neg l \vee l'\} \models \neg l \vee l'$ if and only if $\Pi\backslash\{\neg l \vee l'\} \cup \{l\} \models_{UP} l'$. Therefore, we can consider the graph induced by the unit propagation of $l$ in $\Pi$, and check whether $l'$ is reachable from $l$ without using the edge corresponding to $\neg l \vee l'$. This test can be done in linear time by a modified algorithm of graph reachability.

The exact description of algorithms, and the proofs of their correctness, are given in the following three sections.

## 3.1  Redundancy Checking: Inconsistent 2CNF Formulae

Lemma 4 shows that every inconsistent 2CNF formula contains some clauses allowing both $x$ and $\neg x$ to derive $\bot$ by unit propagation. We show that the number of such clauses is necessarily linear in the number of variables, thus proving that every inconsistent 2CNF formula having a number of clauses that is not linear in the number of variables is redundant.

**Lemma 16** *A 2CNF formula $\Pi$ is inconsistent and irredundant if and only if it is composed of two simple chains of clauses like the following ones:*

$$x \vee l_1, \quad \neg l_1 \vee l_2, \quad \ldots \quad , \quad l_m \vee y, \quad \neg y \vee s_1, \quad \ldots \quad , \quad s_m \vee \neg y$$
$$\neg x \vee p_1, \quad \neg p_1 \vee p_2, \quad \ldots \quad , \quad p_m \vee z, \quad \neg z \vee q_1, \quad \ldots \quad , \quad q_m \vee \neg z$$

*Proof.* If $\Pi$ is inconsistent, by Lemma 4 there exist a variable $x$ such that $\Pi \cup \{x\} \models_{UP} \bot$ and $\Pi \cup \{\neg x\} \models_{UP} \bot$. In turns, $\Pi \cup \{x\} \models_{UP} \bot$ implies the existence of a cycle-less chain allowing the derivation of $y$ and $\neg y$ from $x$ by unit propagation, as explained in Section 2.4. The same holds for $\Pi \cup \{\neg x\} \models_{UP} \bot$. The clauses of these two chains imply inconsistency. Therefore, if $\Pi$ contains other clauses, they are redundant. □

This lemma shows that every inconsistent and irredundant set of clauses is composed exactly of two chains of clauses, each one not containing the same literal twice. The length of each such chain is at most the number of literals; therefore, the number of clauses of such formula is at most two times the number of literals. Therefore, if an inconsistent 2CNF formula contains a number of clauses that is greater than four times the number of its variables, it is redundant. If it contains less clauses, $O(nm)$ and $O(mm)$ are the same. Therefore, checking the consistency of $\Pi \backslash \gamma$ for each $\gamma \in \Pi$ has complexity $O(nm)$.

## 3.2 Redundancy Checking: Consistent 2CNF Formulae Implying Literals

We study the problem of checking redundancy of a set of clauses in which some literals are implied. We show that we can check the redundancy of all clauses $l \vee l'$ such that $\Pi \models l$ in linear time. In other words, time $O(m)$ is required for every literal $l$ that is implied by $\Pi$. The redundancy of the other clauses can be then checking by verifying the redundancy $\Pi \backslash (\Pi | l) \cup \{l\}$ by Lemma 1. After this check has been done for all literals that are entailed by $\Pi$, we obtain a the formula $\Pi \backslash (\Pi | \Pi_{\models}) \cup \Pi_{\models}$ whose parts do not share variables by Lemma 2 and whose first part $\Pi \backslash (\Pi | \Pi_{\models})$ do not entail any literal. The redundancy of the first part can therefore be checked using the algorithm of the next section.

Let $l$ be a literal such that $\Pi \models l$. By Lemma 2, we have $\Pi \cup \{\neg l\} \models_{UP} \bot$, which means that unit propagation from $\neg l$ derives both a literal and its negation. In other words, there exists a variable $x$ such that $\Pi \cup \{\neg l\} \models_{UP} x$ and $\Pi \cup \{\neg l\} \models_{UP} \neg x$. By definition, there are then two acyclic paths in the graph of $\Pi$ induced by $\neg l$, one from $\neg l$ to $x$ and one from $\neg l$ to $\neg x$. The first clause of these two paths are the only clauses that are necessary to allow the derivation of $x$ and $\neg x$ from $\neg l$. Regardless of whether these two clauses are the same or not, they are the only clauses containing $l$ that are necessary to prove $\Pi \cup \{\neg l\} \models_{UP} \bot$. As a result, if $l$ is contained in more than two clauses of $\Pi$, this formula is redundant.

In order to check the redundancy of clauses $l \vee l'$ such that $\Pi \models l$, we first check whether the number of such clauses is greater than two. If this is the case, the set is redundant. Otherwise, we check the redundancy of the clauses $l \vee l'$ by simply performing the linear-time entailment check $\Pi \backslash \{l \vee l'\} \models l \vee l'$ for all such clauses $l \vee l'$. Since there are are most two such clauses, this test only requires linear time. This test is repeated for all literals $l$ such that $\Pi \models l$; therefore, the total running time is $O(nm)$.

## 3.3 Redundancy Checking: Consistent 2CNF Formulae not Implying Literals

We show that the redundancy of 2CNF consistent formulae not implying literals can be checked in time $O(nm)$, where $n$ is the number of literals and $m$ is the number of clauses of the formula.

Let $\Pi$ be a consistent 2CNF formula not implying any literal, and let $\neg l \vee l'$ be one of its clauses. Lemma 3 can be simplified thanks to the assumption that no literal is implied: $\Pi \backslash \{\neg l \vee l'\} \models \neg l \vee l'$ holds if and only if $\Pi \backslash \{\neg l \vee l'\} \cup \{l\} \models_{UP} l'$. Indeed, neither $\neg l$ nor $l'$ are implied by $\Pi$, so they cannot be implied by $\Pi \backslash \{\neg l \vee l'\}$ either.

We therefore only have to check whether $\Pi \backslash \{\neg l \vee l'\} \cup \{l\} \models_{UP} l'$, which can be done by checking whether the graph induced by $l$ on $\Pi$ contains a path from $l$ to $l'$ that does not contain the edge corresponding to the clause $\neg l \vee l'$. We now show that this check can be done for all clauses containing $\neg l$ at the same time in $O(m)$. Let $C_\Pi(l)$ be the following set of literals.

$$C_\Pi(l) = \{l' \mid \neg l \vee l' \in \Pi\}$$

Redundancy of a clause $\neg l \vee l'$ is equivalent to the existence of a path in the graph of $\Pi$ induced by $l$ from another literal in $C_\Pi(l)$ to $l'$. In general, if there is a path from a node in $C_\Pi(l)$ to another node in $C_\Pi(l)$ not containing $l$, the formula is redundant. For example, the following formula is redundant, as we can delete the edge $l \to l''$, and still $l''$ is reachable from $l$.

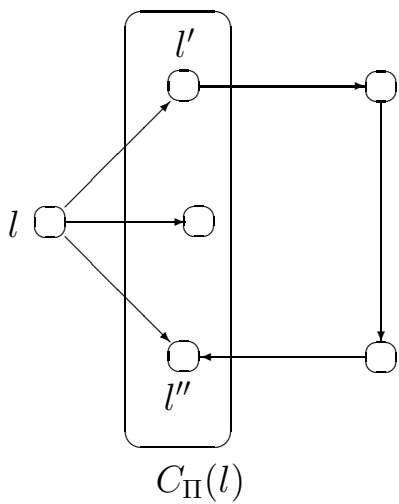

The first step of the algorithm is to remove $l$ and all its incident edges from the graph. The second step is that of checking the existence of a path from $C_\Pi(l)$ to $C_\Pi(l)$ in the resulting graph. Note that no pair of opposite literals can be reached from $l$ because $\Pi$ does not entail the literal $\neg l$.

A variant of the algorithm for node reachability can be used for doing this check while visiting the graph only once. The original algorithm for reachability is as follows.

1. the starting node is marked 0 while all other nodes are marked $\infty$;

2. at each step, take the set $N_H$ of nodes that are marked with the highest integer; let $i$ be this integer;

    (a) for each $n \in N_H$, consider any of its successors $m$;

    (b) mark each $m$ with the minimum among $i+1$ and its previous marker.

3. if no label has been changed during Step 2, stop.

The idea is that the label of the node is the distance from the starting node to it. By visiting the graph width-first, we are considering an edge at most once in the whole process. This is why the algorithm is linear.

This algorithm can be applied to the problem of redundancy if the graph is acyclic: start with the nodes in $C_\Pi(l)$ (instead of a single node), and visit the graph until a node in $C_\Pi(l)$ is reached. In other words, if we reach a situation in which the successor $m$ of a node $n \in N_H$ is in $C_\Pi(l)$, then the set of clauses is redundant.

The algorithm does not work for cyclic graphs: the formula is redundant if a node in $C_\Pi(l)$ can be reached from *another* node in $C_\Pi(l)$. On the contrary, if a node in $C_\Pi(l)$ can be reached from itself, the formula is not necessarily redundant. More precisely, a cycle of this kind does not prove redundancy. In the following example, a node in $C_\Pi(l)$ is reachable from a node in $C_\Pi(l)$, but the path is a cycle; clearly, the set is not redundant.

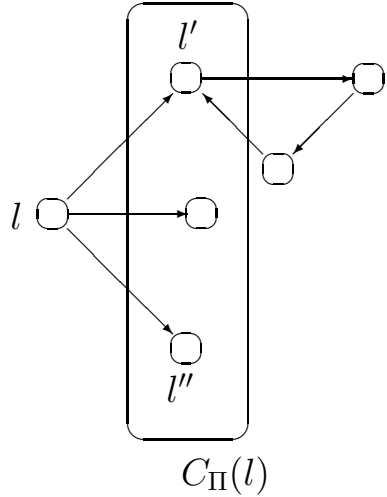

$C_\Pi(l)$

The algorithm must be modified in such a way it checks whether a node of $C_\Pi(l)$ can be reached from *another* node of $C_\Pi(l)$. This can be done by marking each node we visit not only with its distance from $C_\Pi(l)$, but also with the nodes of $C_\Pi(l)$ it is reachable from.

This variant of the node reachability algorithm is however not linear because the set of nodes of $C_\Pi(l)$ a node is reachable from may grow to contain all nodes. However, when a node is reachable from *two* or more different nodes of $C_\Pi(l)$, we do not have to care about the node we started from any longer. Indeed, if a node $l'$ is reachable from two (or more) nodes $l_1, l_2 \in C_\Pi(l)$, and a node of $l_3 \in C_\Pi(l)$ is reachable from $l'$, then either $l_1$ or $l_2$ must be different from $l_3$. As a result, the graph contains a path from $l_1$ or $l_2$ to $l_3$ and at least one between $l_1$ and $l_2$ is different from $l_3$. As a result, once a node $l$ is known to be reachable from two different nodes of $C_\Pi(l)$, we only need to check whether a node of $C_\Pi(l)$ can be reached from $l$.

The algorithm is as follows. In a first phase, labels of nodes can be either $(n, i)$, where $n \in C_\Pi(l)$ and $i$ is an integer, or the special mark two.

1. each node of $n \in C(l)$ is marked with $(0, n)$; take $N_0 = C(l)$; set $i = 0$;

2. set $N_{i+1} = \emptyset$;

3. let $m \in N_i$, and let $(i, n)$ be its marker; for any of its successors $t$:

    (a) if $t$ is not marked, mark it with $(i+1, n)$, and put $t$ in $N_{i+1}$;
    (b) if $t$ is marked with $(j, n)$, it must be $j \leq i$; do not change its marker;
    (c) if $t$ is marked with $(j, s)$, with $n \neq s$, mark it with two;
    (d) if $t$ is marked with two, do not change its mark.

4. if $N_{i+1}$ is empty, stop; otherwise, set $i = i + 1$ and go to step 2.

This is almost the usual visit of the graph width-first. The point is that we mark the nodes not only with their distance from $C_\Pi(l)$, but also with the node they can be reached from. Whenever a node is found out to be reachable from two nodes of $C_\Pi(l)$, we mark it with two and do not continue the search from it. If a node $m \in C_\Pi(l)$ is the successor of a node marked with $(n, i)$, and $n \neq m$, then the graph is redundant. Note that the whole algorithm is linear, as each edge is at most traversed once.

We now have to visit the successors of the nodes we have marked with two, and check whether any node of $C_\Pi(l)$ can be reached from them. This can be done with the very same original reachability algorithm, which is still linear in time.

This algorithm determines the redundancy of all clauses containing one literal in time $O(m)$. Therefore, all clauses can be checked in time $O(nm)$.

# 4 Irredundant Equivalent Subsets (I.E.S.'s)

The following problems about I.E.S.'s are polynomial for 2CNF formulae because of the polynomiality of entailment for this restriction.

**check whether a formula is an I.E.S. of another formula.** Check containment and irredundancy;

**check whether a clause is in all I.E.S.'s.** A clause $\gamma$ is in all I.E.S.'s of a formula $\Pi$ if and only if $\Pi \backslash \{\gamma\} \not\models \gamma$ [Lib05];

**uniqueness.** A formula $\Pi$ has an unique I.E.S. if and only if $\{\gamma \in \Pi \mid \Pi \backslash \{\gamma\} \not\models \gamma\} \models \Pi$ [Lib05].

Two other problems require a more detailed analysis: the size of I.E.S. (checking whether a formula has an I.E.S. of size bounded by a number $k$) and the presence in an I.E.S. (checking whether a given clause is present in some I.E.S. of a given formula). The presence of cycles of clauses in the formula mostly determine the complexity of these two problems.

As it is clear from the two tables below, the complexity (at least in the cases that have been successfully classified) does not depend on whether the formula is consistent or implies literals. However, the proofs are different in the various cases. In the two tables below, "single" indicates formulae implying a single literal and "nonsingle" indicates formulae not implying any literal.

**Size of I.E.S.**

|         | inconsistent | single  | nonsingle |
|---------|--------------|---------|-----------|
| acyclic | P            | P       | P         |
| cyclic  | ??           | NP-hard | NP-hard   |

**Presence in a I.E.S.**

|         | inconsistent | single  | nonsingle |
|---------|--------------|---------|-----------|
| acyclic | P            | P       | P         |
| cyclic  | NP-hard      | NP-hard | NP-hard   |

We use the following order of the cases: first, all acyclic cases, then all the cyclic cases; in each case, we first consider inconsistent formulae, then consistent formulae implying literals, and then consistent formulae not implying literals.

The following two results about consistent acyclic 2CNF formulae have already been proved.

- Acyclic consistent formulae not implying single literals always have a unique I.E.S.

- Bulding an I.E.S. for an acyclic consistent formula implying the literals $\neg l_1, \ldots, \neg l_n$ can be done by choosing independently some clauses containing $\neg l_1$, some clauses containing $\neg l_2$, etc.

We show that these two fact allow proving proving proving proving proving proving proving proving proving that, for acyclic consistent 2CNF formulae, one can determine presence and size of I.E.S.'s in polynomial time.

**Theorem 2** *The unique I.E.S. of a consistent acyclic 2CNF formula not implying literals can be found in polynomial time.*

*Proof.* Let $\Pi$ be a consistent acyclic 2CNF formula. Since it has a single I.E.S., a clause is in its I.E.S. if and only if it is in all its I.E.S.'s. Since checking presence can be done by checking $\Pi \backslash \{\gamma\} \models \gamma$, and this inference is linear-time for 2CNF formulae, we can conclude that checking the presence of each clause in the unique I.E.S. of the formula can be done in linear time. □

The proof of the theorem shows a quadratic algorithm for generating the single I.E.S. of $\Pi$ based only one the uniqueness of such an I.E.S. However, Lemma 4 allows for a slightly better algorithm: for each $l$, can check all clauses $\neg l \vee l_1$ at once by visiting the graph $R_\Pi(l)$. Since the construction and visit of $R_\Pi(l)$ takes linear time, the overall running time is $O(nm)$, where $n$ is the number of variables and $m$ is the number of clauses.

**Theorem 3** *The problems of presence of a clause in an I.E.S. and of existence of an I.E.S. of given size are polynomial-time for acyclic consistent 2CNF formulae.*

*Proof.* By Lemma 15, the presence of a clause $l \vee l'$ with $\Pi \models \neg l$ in an I.E.S. only depends on whether $l' \in S_\Pi(l)$ or there exists $l''$ such that $(l', l'') \in P_\Pi(l)$. Since these sets $S_\Pi(l)$ can be checked in polynomial time by definition, checking the presence of a clause $l \vee l'$ in an I.E.S. is a polynomial-time problem if $\Pi \models \neg l$. All other clauses can be checked in polynomial time thanks to Corollary 3 and Theorem 2.

One of the smallest I.E.S.'s of an acyclic formula can be built in a similar way but choosing $\neg l$ or a clause $\neg l \vee l'$ such that $l' \in S_\Pi(l)$ if possible, and a pair of clauses $\neg l \vee l'$ and $\neg l \vee l''$ with $(l', l'') \in P_\Pi(l)$ otherwise. To keep into account the pair of clauses $\neg l \vee l'$ and $\neg l \vee \neg l'$ that replace the unit clause $l$ due to the transformation of Section 2.4, we count such a pair as it were a single clause of $\neg l \vee l'$ with $l' \in S_\Pi(l)$. □

# 5 Size of I.E.S.

In this section, we consider the problem of checking whether a formula has an I.E.S. of size bounded by a given integer $k$. This problem has already been proved polynomial for consistent acyclic 2CNF formulae. The remaining cases are:

- acyclic inconsistent 2CNF formulae;
- cyclic 2CNF formulae, in all three cases.

## 5.1 Size of I.E.S.: Acyclic Inconsistent 2CNF Formulae

Let $\Pi$ be an inconsistent and acyclic 2CNF formula. By Lemma 4, $\Pi$ is inconsistent if and only if there exists a variable $x$ such that $\Pi \cup \{x\} \models_{UP} \bot$ and $\Pi \cup \{\neg x\} \models_{UP} \bot$. As explained in Section 2.4, $\Pi \cup \{x\} \models_{UP} \bot$ implies that there exists a path from $x$ to $l_1$ and from $l_1$ to $\neg l_1$. For the same reason, $\Pi \cup \{\neg x\} \models_{UP} \bot$ implies that there is a path from $\neg x$ to $l_2$ and from $l_2$ to $\neg l_2$. These paths can share nodes. We consider the various possible cases.

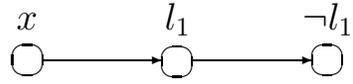

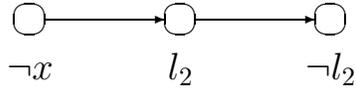

Case 1: disjoint paths.

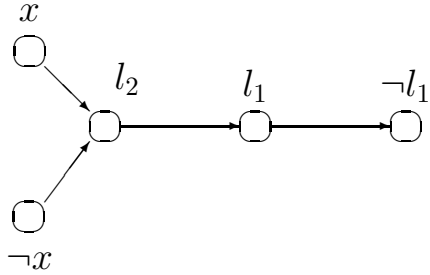

Case 2: a common part "before the inconsistency".

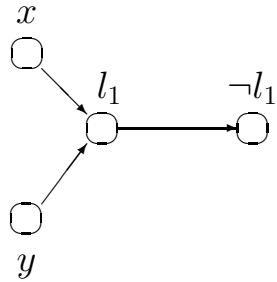

Case 3: a common part "at the inconsistency".

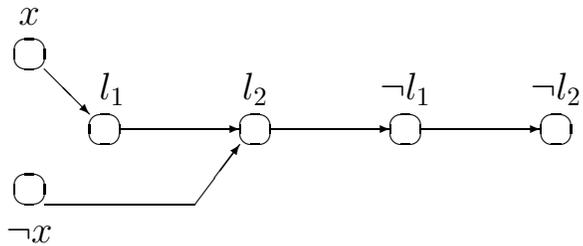

This formula is cyclic: we can go from $\neg x$ to $\neg l_1$ and from $\neg l_1$ to $\neg x$ by inverting the path from $x$ to $l_1$.

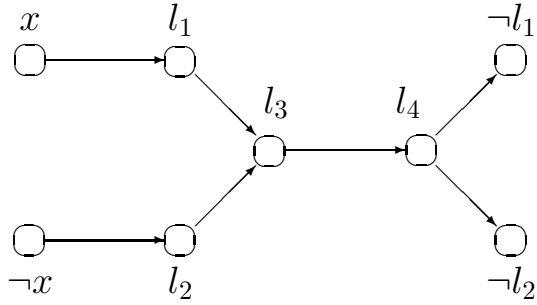

This formula is cyclic: we can go from $x$ to $\neg l_2$, and from $\neg l_2$ to $x$ by inverting the path from $\neg x$ to $l_2$.

We can check the number of edges needed to reach each literal from each other: we build a table containing the number of clauses needed to go from any literal to any other one, if possible. We can then consider each case separately, and calculate how many clauses are necessary to form one of the patterns above.

For example, for checking case 1, we consider all possible triple of literals $\langle x, l_1, l_2 \rangle$, and sum up the distance from $x$ to $l_1$, the distance from $l_1$ to $\neg l_1$, etc. For each triple, we obtain the number of clauses needed to reach inconsistency according to Case 1. This procedure is done also for the other cases, and the minimal number is selected.

This algorithm is correct if the formula does not contain cycles. If the formula contains cycles, the algorithm is incorrect because it does not take into account that two edges in the above graphs can correspond to the same clause and should therefore be counter once, not twice.

We prove that this is not possible if the formula is acyclic. Let $a \vee b$ be a clause. Its associated edges are $\neg a \rightarrow b$ and $\neg b \rightarrow a$. This clause is "counted twice" if:

1. the same edge occur twice in the same path: if $\neg a \rightarrow b$ occurs twice in the same path, we have a cycle from $\neg a$ to $\neg a$;

2. the same edge occur once in both paths but not in their common part (if any): if $\neg a \rightarrow b$ occurs both in the path from $x$ and in the path from $\neg x$, we are in the situation in which $\neg a$ is reachable from both $x$ and $\neg x$ and from $\neg a$ we can reach a pair of opposite literals; the paths in which what follows $\neg a$ is only counted once has been then considered as a Case 2 or Case 3 in which $\neg a$ is the point where the paths join;

3. the edge $\neg a \rightarrow b$ and $\neg b \rightarrow a$ occur in the same path: we can remove the last edge and what follows because contradiction has already been reached;

4. the edge $\neg a \rightarrow b$ and $\neg b \rightarrow a$ occur in two different paths: for example, if the first clause occurs in the path from $x$ and the second in the path from $\neg x$, we can go from $x$ to $\neg a$ to $b$, and then from $\neg b$ to $\neg x$ by reversing the path from $\neg x$ to $b$.

We can conclude that, if we check all possible cases, we always end up with the minimal ones in which each edge corresponds to a unique clause. Since checking distances in graphs is linear, we can prove the following theorem.

**Theorem 4** *Checking whether an acyclic inconsistent 2CNF formula has an inconsistent subset of size less that or equal to k is polynomial.*

*Proof.* We determine the minimal number of edges that are necessary to form any of the four combinations above. We have proved that any way of reaching inconsistency can be recast as one of them with no repeated clause. Therefore, the minimal number of edges of any case is the minimal number of clauses needed to form an inconsistency. □

In order to keep into account the fact that a pair of clauses $l \vee l'$ and $l \vee \neg l'$ might represent the same original unit clause $l$ because of the transformation of Section 2.4, we count 1/2 instead of 1 the edges that have been introduced by this transformation.

## 5.2 Size of I.E.S.: Cyclic Consistent 2CNF Formulae Implying Literals

We show that Lemma 15 does not hold if the literal $l$ under consideration is in a cycle of clauses. Consider the following formula:

$$\Pi = \{\neg l' \vee l, \neg l \vee l', \neg l \vee l'', \neg l' \vee x, \neg l' \vee \neg x, \neg l'' \vee y, \neg l'' \vee \neg y\}$$

The graph of $\Pi$ induced by $l$ is the following one, showing a simple cycle between the literals $l$ and $l'$.

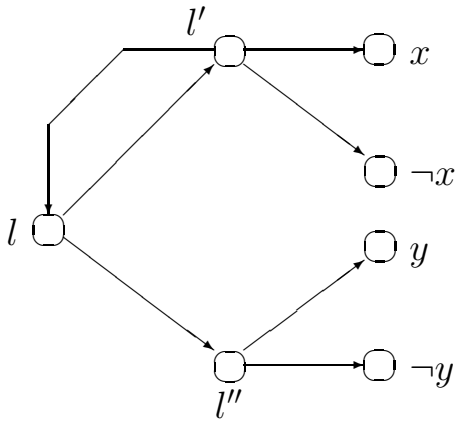

For this formula, $S_\Pi(l) = \{l', l''\}$, while $P_\Pi(l) = \emptyset$. According to Lemma 15, if $\Pi'$ is an I.E.S. of $\Pi$, then we can replace its clauses of $D_\Pi(l)$ with a single clause $S_\Pi(l)$ and what results is still an I.E.S. of $\Pi$.

We show a counterexample. The formula $\Pi$ is equivalent to $\{\neg l, \neg l', \neg l''\}$. The following is an I.E.S. of $\Pi$.

$$\Pi' = \{\neg l' \vee l, \neg l \vee l'', \neg l'' \vee y, \neg l'' \vee \neg y\}$$

Graphically, $\Pi'$ is the following subset of the formula:

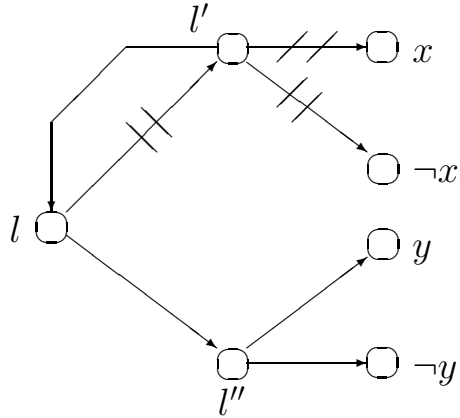

This subset is equivalent to $\Pi$ because it entails all three literals $l$, $l'$, and $l''$. Indeed, $l''$ is entailed by the clauses $\neg l'' \vee y$ and $\neg l'' \vee \neg y$, while $l$ and $l'$ are entailed thanks to $\neg l' \vee l$ and $\neg l \vee l''$, which make $l''$ reachable by unit propagation from $l$ and $l'$.

According to Lemma 15, we should be able to obtain another I.E.S. from $\Pi'$ by replacing all clauses of $D_\Pi(l)$ with $\neg l \vee l'$ in it, because $l' \in S_\Pi(l)$. Let $\Pi'' = \Pi' \backslash D_\Pi(l) = \{\neg l' \vee l, \neg l'' \vee y, \neg l'' \vee \neg y\}$. If Lemma 15 were true for cyclic formulae, it would be that $\Pi'' \cup \{\neg l \vee l'\}$ is another I.E.S. This is however false, as this set does not imply neither $\neg l$ nor $\neg l'$. Graphically, $\Pi'' \cup \{\neg l \vee l'\}$ is the following formula:

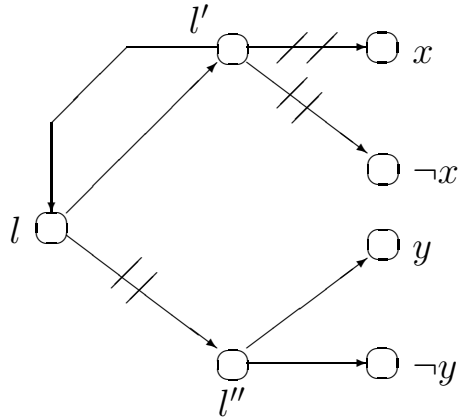

This formula still entails $\neg l''$, but it does no more entail neither $\neg l$ nor $\neg l'$. These are indeed the two literals involved in the only cycle of this formula. Intuitively, Lemma 15 does not hold for cyclic formulae because, while $\Pi' \cup \{l'\} \models_{UP} \bot$ still hold in any I.E.S. $\Pi'$ of $\Pi$ for every $l' \in S_\Pi(l)$, the unit derivation of $\bot$ from $l'$ might involve the literal $l$. Therefore, the choice of clauses containing $l$ cannot be done independently of the choices of the other clauses.

We now consider a generalization of the example above: we are given a set of literals such that:

1. their negations are all entailed by $\Pi$;

2. each literal in the set is reachable from any other one.

We study the problem of finding a minimal set of clauses containing these literals and that are part of an I.E.S. Given any literal $l$ such that $\Pi \models \neg l$, its induced graph is composed of a strongly connected component including $l$, joined to the rest of the graph. Regarding the rest of the graph, we are only interested in the nodes that are joined to a node of the component.

For every literal $l$, we define $CC_\Pi(l)$ as the following set of literals.

NOTATION: $CC_\Pi(l) = \{l' \mid l \text{ and } l' \text{ are in a cycle in the graph induced by } l \text{ on } \Pi\}$

Since $\Pi \models l'$ for every $l' \in CC_\Pi(l)$, the same holds in every I.E.S. of $\Pi$. Assuming that $\Pi$ contains no unary clause as discussed in Section 2.4, the graph of $\Pi$ induced by $l$ is composed of a strongly connected component made of the nodes of $CC_\Pi(l)$, and other nodes and edges. Since $\Pi$ implies all literals of $CC_\Pi(l)$, this graph also contains paths from nodes of $CC_\Pi(l)$ to pairs of opposite literals. These literals cannot both be part of $CC_\Pi(l)$ as otherwise $\Pi$ would be inconsistent.

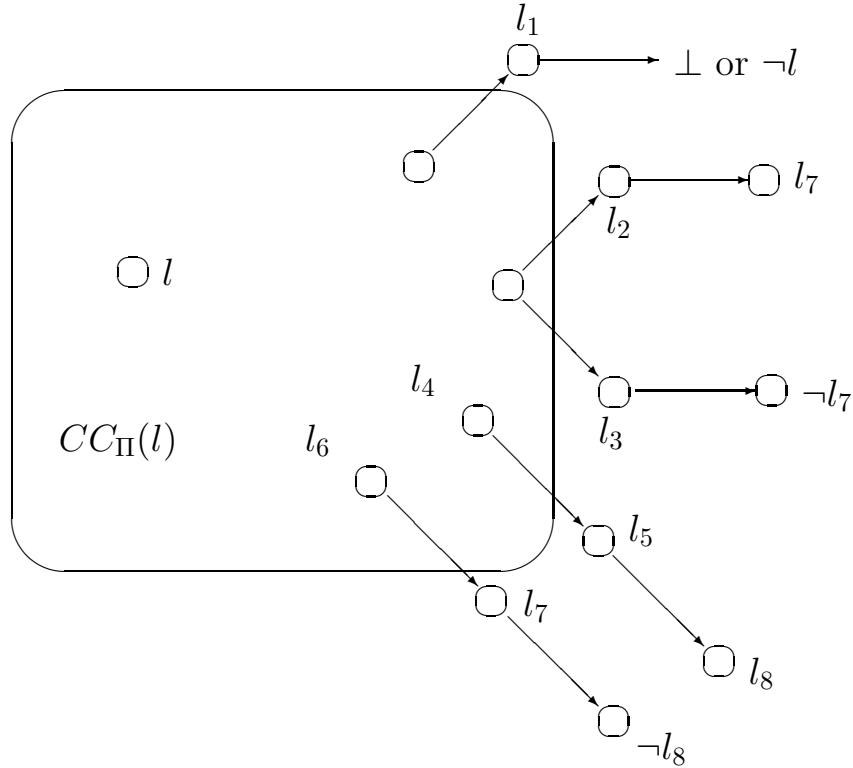

This figure shows the three possibilities: either from a node of $CC_\Pi(l)$ we can derive $\bot$, or from two nodes of $CC_\Pi(l)$ we can reach a pair of inconsistent literals. The nodes that are reachable with a single edge from $CC_\Pi(l)$ will be denoted by $JC_\Pi(l)$:

NOTATION: $$JC_\Pi(l) = \{l' \mid \exists l \in CC_\Pi(l) \text{ and } \neg l \vee l' \in \Pi\}$$

The idea is that we do not need to care about "what happens outside $CC_\Pi(l)$": if $\bot$ or $\neg l$ can be reached from a literal $l_1$ of $JC_\Pi(l)$ in $\Pi$, that will also hold for every I.E.S. of $\Pi$. Since $CC_\Pi(l)$ is the largest connected componente including $l$, this path from $l_1$ to $\bot$ or $\neg l$ cannot re-enter $CC_\Pi(l)$. Therefore, which clauses are chosen from $CC_\Pi(l)$ to be in the I.E.S., and which clauses are chosen in the outside of $CC_\Pi(l)$ do not interact with each other.

For each $l' \in JC_\Pi(l)$ we define $LC_\Pi(l')$ to be the set of literals that can be derived from $l$ using unit propagation. Since $l'$ is not in $CC_\Pi(l)$, unit propagation cannot include literals in $CC_\Pi(l)$.

NOTATION: $$LC_\Pi(l) = \{l' \mid \exists l \in JC_\Pi(l) \text{ and } \Pi \cup \{\neg l\} \models_{UP} l'\}$$

The idea is that all literals of $CC_\Pi(l)$ imply $\bot$ in $\Pi$, and the same must therefore happen in any I.E.S. of $\Pi$. Moreover, any I.E.S. of $\Pi$ must allow deriving all literals of $LC_\Pi(l')$ from $l'$, and the clauses involved in this unit propagation cannot contain literals in $CC_\Pi(l)$. Therefore, we can choose the clauses in $CC_\Pi(l)$ assuming that each $l' \in JC_\Pi(l)$ implies $LC_\Pi(l')$ regardless of this choice.

In the figure above, the nodes in $CC_\Pi(l)$ are all connected to each other. The nodes $l_1, \ldots, l_6$ form $JC_\Pi(l)$, which is the "frontier" of the rest of the graph. Since $l$ implies $\bot$, $LC_\Pi(l)$ contains either a node like $l_1$, which implies $\bot$ or $\neg l$ alone, or a pair of nodes like $l_2$ and $l_3$, that imply a pair of opposite literals. Any such pair may be reachable either from a single node of $CC_\Pi(l)$, like $l_2$ and $l_3$, or from two different nodes, like $l_4$ and $l_5$.

The idea is that any I.E.S. will allow deriving $\bot$ or $\neg l$ from $l_1$ without using any clause with literals in $CC_\Pi(l)$, as otherwise $l_1$ would be in $CC_\Pi(l)$ as well. For the same reason, it will be possible to derive $l_7$ from $l_2$, etc. Therefore, we can choose the subset of clauses with literals in $CC_\Pi(l)$ first, and then complete the I.E.S. by independently selecting literals in the rest of the graph. Therefore, once we have chosen clauses with one literal in $CC_\Pi(l)$, the problem reduces to the acyclic case.

How can we choose a subset of edges joining nodes of $CC_\Pi(l)$ to form a minimal I.E.S.? In this subset, every literal in $CC_\Pi(l)$ must imply $\bot$. If a subset has this property, all other clauses whose edge starts from a node of $CC_\Pi(l)$ are redundant. We consider three cases separately, and establish the size of minimal I.E.S. in each case.

$JC_\Pi(l)$ **contains a literal that implies $\bot$ or $\neg l$ in $\Pi$.** In the example above, $l_1$ is in this condition. We prove that the minimal size of I.E.S.'s is equal to the number of literals of $CC_\Pi(l)$.

Choose a node that is connected to $l_1$, and add the edge from this node to $l_1$, and proceed recursively adding edges only for nodes that are not already connected to $l_1$. This algorithm makes all nodes of $CC_\Pi(l)$ connected to $l_1$ because $CC_\Pi(l)$ is strongly connected: indeed, each node is visited, and an edge is added to connect it to $l_1$ if it is not already. Moreover, for each node the resulting set of edges contains at most one outgoing edge. This proves that the I.E.S. has size equal to that of $CC_\Pi(l)$. No I.E.S.

can be smaller: a smaller set of edges necessarily leaves one node not joined to any edge, thus making it not implying $\bot$ as required.

**A node of $CC_\Pi(l)$ is connected to two nodes of $JC_\Pi(l)$ that imply $\bot$.** In the example above, $l_2$ and $l_3$ are two nodes in this condition. We assume that no node of $JC_\Pi(l)$ implies $\bot$ or $\neg l$, that is, we are not in the case above.

We prove that all nodes of $CC_\Pi(l)$ can be made connected to $\bot$ by choosing a number of edges that is equal to the size of $CC_\Pi(l)$ plus one. Such a set of edges exists because all nodes can be connected to $\bot$ by selecting the two edges to $l_2$ and to $l_3$ and then repeating the algorithm of the previous case from the common predecessor.

We now prove that no smaller subset of edges makes all nodes of $CC_\Pi(l)$ connected to $\bot$. Let us assume the converse: there exists a subset of $|CC_\Pi(l)|$ edges that makes all nodes of $CC_\Pi(l)$ connected to $\bot$. Since each node has to be connected to some other node, we have that each node $l'$ of $CC_\Pi(l)$ is connected to exactly one other node $l''$. This other node can be either in $CC_\Pi(l)$ or in $JC_\Pi(l)$. In the second case, $l'$ is only connected to $l''$. In order for $l'$ to be connected to $\bot$, then $l''$ must be connected to $\bot$ or $\neg l$ as well, contradicting the assumption that no node of $JC_\Pi(l)$ is connected to $\bot$ or $\neg l$. We can therefore conclude that any node of $CC_\Pi(l)$ is connected to exactly one other node of $CC_\Pi(l)$. However, $CC_\Pi(l)$ cannot contain a pair of opposite literals, as otherwise $\Pi$ would be inconsistent. As a result, no node of $CC_\Pi(l)$ is connected to $\bot$, contradicting an assumption.

**Two nodes of $CC_\Pi(l)$ are joined to two nodes of $JC_\Pi(l)$ implying $\bot$.** Here, we assume that we are not in one of the two cases above: no single literal of $JC_\Pi(l)$ is connected to $\bot$, and no single literal of $CC_\Pi(l)$ is connected to two literals of $JC_\Pi(l)$ implying two opposite literals. In the example above, the two nodes $l_5$ and $l_7$ of $JC_\Pi(l)$ are in the conditions we assumed. We are also assuming that no pair of nodes are like $l_2$ and $l_3$, and that no single node is like $l_1$.

Let us assume there is only one pair of nodes in the conditions we assumed. We show how subset of minimal size connecting all nodes of $CC_\Pi(l)$ to $\bot$ are made. First, take the edge from $l_4$ to $l_5$ and from $l_6$ to $l_7$. If there exists a simple cycle containing both $l_4$ and $l_6$, take its edges, and then visit the rest of $CC_\Pi(l)$ backwards by adding edges from nodes that are not already connected to both $l_4$ and $l_6$. This set of edges is composed of $|CC_\Pi(l)| + 1$ edges, and no smaller set maintains all nodes of $CC_\Pi(l)$ connected to $\bot$.

The point is that $l_4$ has to be connected to $l_6$ and vice versa. Therefore, any I.E.S. contains a cycle including $l_4$ and $l_6$. Simple cycles have exactly one edge for each node, and are therefore optimal from this point of view, as all nodes of the cycle have to be connected to $l_4$ and $l_6$ anyway.

If there is no simple cycle including both $l_4$ and $l_6$, we consider a cycle composed of more than one simple cycle. If two cycles have one or more common points, we have one more edge w.r.t. the optimal situation. Therefore, we have to minimize the number of common points among cycles.

In the first two cases, finding the minimal size of I.E.S.'s is easy, as it amounts to counting the number of nodes in $CC_\Pi(l)$, and adding one in the second case. In the third case, however, we have to check the existence of a simple cycle including two nodes. This problem is NP complete.

**Theorem 5** *Deciding whether a graph contains a simple cycle including two given nodes is* NP*-complete.*

*Proof.* Membership is ovious via a guess-and-check algorithm. Hardness is probed by reduction from the problem path via a node. This is the problem of determining whether there exists a simple path from a node $x$ to a node $y$ that includes a node $m$, in graph $G$. This problem is NP complete [LP84].

Given a graph $G$ and three nodes $x$, $y$, and $m$, we build a graph $G'$ by first removing all edges that are incoming to $x$, and then adding the edge from $y$ to $x$. This graph contains a simple cycle including both $x$ and $m$ if and only if there exists a simple path from $x$ to $y$ including $m$.

First, the removal of incoming edges to $x$ does not change the simple paths from $x$ to $y$, as no such path, being simple, contains an edge that is incoming to $x$. For the same reason, the addition of the edge from $y$ to $x$ does not change the set of simple paths from $x$ to $y$. However, since the edge from $y$ to $x$ is the only incoming edge to $x$, it is contained in any cycle including $x$. Therefore, any simple cycle including $x$ is composed of a simple path from $x$ to $y$ and of the edge from $y$ to $x$. Therefore, a path from $x$ to $y$ including $m$ exists if and only if a simple cycle including both $x$ and $m$ exists. □

This theorem shows that the problem of checking whether a set of clauses implying a single literal, and containing cycles in the graph induced by the negation of this literal, is NP-complete.

**Theorem 6** *Deciding whether a consistent 2CNF formula implying a single literal $l$ such that the graph induced by $\neg l$ on the graph contains cycles has a* I.E.S. *of size $k$ is* NP*-complete.*

*Proof.* Hardness is proved by reduction from the problem of checking the existence of a simple cycle in a graph containing two given nodes $x$ and $y$. First, check whether $y$ is reachable from $x$ and vice versa: if not, there is no simple cycle including both $x$ and $y$. The second step is the deletion of any node such that $x$ is not reachable from. This step clearly does not alter the set of cycles including $x$. Then, build the set of clauses obtained from the edges of the graph, and then two clauses $\neg x \vee z$ and $\neg y \vee \neg z$. All literals imply $\bot$ because all nodes are connected to $x$ which is connected to $y$. Set $k$ to the number of nodes of the graph plus one. The original graph has a simple cycle including both $x$ and $y$ if and only if we can build an I.E.S. for the set of clauses that is made of a simple cycle including $x$ and $y$, from the clauses $\neg x \vee z$, and $\neg y \vee \neg z$, and one clause for any other literal. □

## 5.3 Size of I.E.S.: Cyclic Consistent Formulae not Implying Literals

The problem of determining whether a formula has an I.E.S. of size at most $k$ is NP-complete if the formula is consistent and cyclic. This is in particular true even if the formula does not entail any literal.

**Theorem 7** *Deciding whether a set of binary clauses $\Pi$ has an equivalent subset of size bounded by $k$ is NP complete. This result holds for sets of clauses that implies that all variables are equivalent and does not entail any single literal.*

*Proof.* The problem of finding a minimum equivalent subgraph is NP-complete. In particular, it remains complete even if the graph is strongly connected [Sah74, KRF95]. This is the problem of finding the minimum number of edges of the graph that makes the resulting graph strongly connected.

We show a reduction from this problem to that of finding whether there exists a I.E.S. of a set $\Pi$ of size bounded by $k$. Let $G = (N, E)$ be a graph. The set of variables of $\Pi$ is the set of nodes $N$. For each edge $(i, j)$ the set $\Pi$ contains the clause $\neg i \vee j$. For any two nodes $i$ and $j$, the node $j$ is reachable from $i$ if and only if $\Pi \cup \{i\} \models_{UP} j$. In order to complete the proof, we only have to show that $\Pi \not\models l$ for any literal $l$: if this is true, then reachability is in one-to-one correspondence with entailment, thus showing that equivalence of graphs implies equivalence of the corresponding formulae.

Since each clause contains a positive and a negative literal, it is satisfied both by the model setting all variables to true and by the model setting all variables to false. These are therefore both models of $\Pi$. However, if one of them satisfies $l$, the other one does not. Therefore, there is a model of $\Pi$ that does not satisfy $l$, thus showing that $l$ is not implied by $\Pi$. □

Note that the set of clauses we used in the proof makes all variables equivalent. This is necessary, as the proof of hardness of the minimum equivalent subgraph holds relies on the graph being strongly connected.

## 6 Presence in an I.E.S.

In this section, we study the problem of checking whether a clause is contained in some I.E.S.'s of a given 2CNF formula. The cases of consistent acyclic sets have already been considered, and the problem proved polynomial in these cases. This result can be slightly extended to the case in which some cycles are present, but none include the clause to check. We then consider the problem of acyclic inconsistent. For consistent cyclic formulae, we show that the problem is NP-complete regardless of whether some literals are implied by the formula. The problem has the same complexity if the set if inconsistent.

## 6.1 Presence in an I.E.S.: Acyclic Inconsistent 2CNF Formulae

In this section, we study the problem of telling whether a clause is in an I.E.S. of a cyclic inconsistent formula. We prove that this problem is polynomial.

In the particular case when $\Pi$ is inconsistent, an I.E.S. of $\Pi$ is an irredundant inconsistent subset of $\Pi$. A clause is in some I.E.S. of an inconsistent formula if there is an inconsistent subset of $\Pi$ that contains this clause, and that becomes consistent if this clause is removed from it. Formally, $l_1 \vee l_2 \in \Pi$ is in some I.E.S. of $\Pi$ if and only if there exists $\Pi' \subseteq \Pi \backslash \{l_1 \vee l_2\}$ such that $\Pi' \not\models \bot$ but $\Pi' \cup \{l_1 \vee l_2\} \models \bot$.

If $\Pi \backslash \{l_1 \vee l_2\}$ is consistent, then $\Pi \backslash \{l_1 \vee l_2\} \not\models l_1 \vee l_2$, and $\gamma$ is therefore in all I.E.S.'s of $\Pi$. From now on, we only consider formulae $\Pi$ such that $\Pi \backslash \{l_1 \vee l_2\}$ is inconsistent. A necessary condition to ensure the presence of the clause $l_1 \vee l_2$ in an I.E.S. of $\Pi$ in this case is given as follows.

**Lemma 17** *If $l_1 \vee l_2$ is in an I.E.S. of an inconsistent acyclic 2CNF formula $\Pi$ such that $\Pi \backslash \{l_1 \vee l_2\}$ is inconsistent, then:*

$$\Pi \backslash \{l_1 \vee l_2\} \cup \{l_1\} \models_{UP} \bot$$
$$\Pi \backslash \{l_1 \vee l_2\} \cup \{l_2\} \models_{UP} \bot$$

*Proof.* We assume that the first equation above is false and prove that $l_1 \vee l_2$ in not in an I.E.S. of $\Pi$. A similar proof can be used assuming that the second equation is false.

Let $\Pi'$ be a consistent subset of $\Pi \backslash \{l_1 \vee l_2\}$. Since $\Pi \backslash \{l_1 \vee l_2\} \cup \{l_1\} \not\models_{UP} \bot$, we have that $\Pi' \cup \{l_1\} \not\models_{UP} \bot$. As a result, $\Pi'$ is consistent with $l_1$ and therefore it is also consistent with $l_1 \vee l_2$. In the other way around, if $\Pi''$ is an inconsistent subset of $\Pi$, then $\Pi'' \backslash \{l_1 \vee l_2\}$ is inconsistent as well. As a result, no I.E.S. of $\Pi$ contains $l_1 \vee l_2$. □

The converse of this lemma is not true, as shown by the following formula.

$$\Pi = \{l_1 \vee l_2, \neg l_1 \vee l_3, \neg l_3 \vee x, \neg l_3, \vee \neg x, \neg l_2 \vee \neg l_3, l_3 \vee y, l_3 \vee \neg y\}$$

The graphs of this formula induced by $l_1$ and $l_2$ are as follows; clearly, $\bot$ is reachable from both $l_1$ and $l_2$ in $\Pi$.

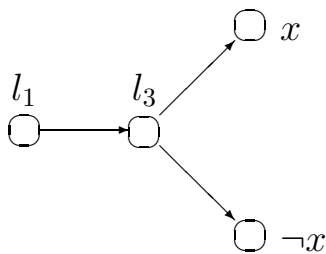 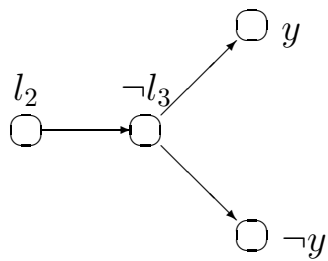

Graph of $\Pi$ induced by $l_1$.    Graph of $\Pi$ induced by $l_2$.

While $\perp$ is reachable from both $l_1$ and $l_2$ in $\Pi$, yet $l_1 \vee l_2$ is not necessary to produce inconsistency in any subset of $\Pi$. This is because any subset of $\Pi$ allowing $\perp$ to be reached from $l_1$ and $l_2$ also allows $\perp$ to be reachable from $l_3$ and $\neg l_3$ without using the clause $l_1 \vee l_2$.

This formula is cyclic: a simple cycle of clauses is $\{l_1 \vee l_2, \neg l_2 \vee \neg l_3, l_3 \vee \neg l_1\}$. We indeed prove that the lemma above can be turned into an "if and only if" whenever $\Pi$ is an acyclic formula.

**Lemma 18** *The clause $l_1 \vee l_2$ is in some I.E.S. of the acyclic inconsistent 2CNF formula $\Pi$ if the following conditions hold: $\Pi \backslash \{l_1 \vee l_2\} \models \perp$, $\Pi \backslash \{l_1 \vee l_2\} \cup \{l_1\} \models_{UP} \perp$, and $\Pi \backslash \{l_1 \vee l_2\} \cup \{l_2\} \models_{UP} \perp$.*

*Proof.* Since $\Pi \backslash \{l_1 \vee l_2\} \cup \{l_1\} \models_{UP} \perp$, there exists a path from $l_1$ to a pair of opposite literals $l_3$ and $\neg l_3$ in $\Pi \backslash \{l_1 \vee l_2\}$, and the same holds for $l_2$. Let $\Pi'$ be the set composed of exactly all clauses used in the unit propagation from $l_1$ to $\perp$ and from $l_2$ to $\perp$. We have that $\Pi' \models \{\neg l_1, \neg l_2\}$. Therefore, $\Pi' \cup \{l_1 \vee l_2\}$ is inconsistent. We prove that $\Pi'$ is consistent, thus proving that $l_1 \vee l_2$ is in all I.E.S.'s of $\Pi'$ and therefore in some I.E.S.'s of $\Pi$.

The paths from $l_1$ and $l_2$ to $\perp$ can be visualized as follows:

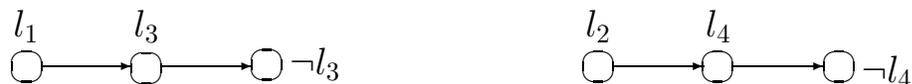

The idea is that we can set to false all literals of these two paths besides the final one of each one, and this would be a model of $\Pi$. However, these two paths can share variables; this assigment might not be a model in this case.

These two paths do not contain opposite literals besides $l_3$ and $l_4$. Indeed, if the path from $l_1$ contains a pair of opposite literals besides $l_3, \neg l_3$, then contradiction is reached before the end of the path. The same holds for the path from $l_2$. If the paths from $l_1$ contains $l_5$ and the path from $l_2$ contains $\neg l_5$, then from $l_1$ we can reach $l_5$ and from $l_5$ we can reach $\neg l_2$; together with the clause $l_1 \vee l_2$, we would have a cycle while $\Pi$ is assumed acyclic.

As a result, these two paths can share literals and clauses, but cannot contain a pair of opposite literals. A satisfying truth assignment is obtained as follows:

1. set all literals to false;

2. set the final literal of each path to true;

3. for every node that is set to true, set its successor (if any) to true.

The third point is necessary because the two paths can share nodes. This assigment sets two opposite literals to true only if both $l_4$ and $\neg l_4$ are reachable from $\neg l_3$ or both $l_3$ and $\neg l_3$ are reachable from $\neg l_4$. We show that the first situation is impossible; a similar proof holds for the second situation.

Since $l_4$ and $\neg l_4$ are reachable from $\neg l_3$, we are in the following situation:

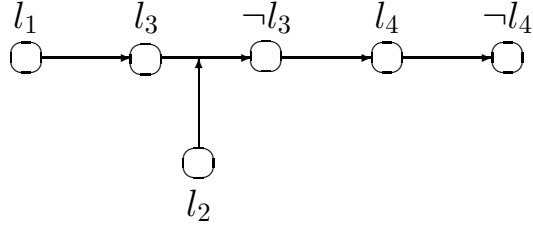

This situation is impossible because it implies the existence of a cycle in $\Pi$: from $l_2$ we can go to $\neg l_3$, and from $\neg l_3$ we can go to $\neg l_1$. The clause $l_1 \vee l_2$ closes the cycle. $\square$

The proof of the theorem above does not hold for cyclic formulae because a literal can be reachable from $l_1$ while its negation is reachable from $l_2$. As a result, the conditions that both $l_1$ and $l_2$ are connected to $\bot$ are not sufficient: we have also to check that there is a set of edges that connect them to $\bot$ while no literal is such that both it and its negation reach $\bot$. In fact, it can be proved that the problem is NP-hard for cyclic formulae.

## 6.2 Presence in an I.E.S.: Acyclic Consistent 2CNF Formulae not Implying Literals

We have already shown that an acyclic consistent 2CNF formula not implying a single literal have a unique I.E.S., which can be determined in polynomial time. Therefore, checking the presence of a clause in an I.E.S. is easy. In this section we extend this result to the case in which the formula contains some cycles, but none include the clause under consideration.

The following figure explains the concept: if we box all literals that are equivalent to $l_1$, and all literals that are equivalent to $l_2$, then $l_1 \rightarrow l_2$ is in some I.E.S. if and only if all paths from one box to the other one only contains nodes in the boxes.

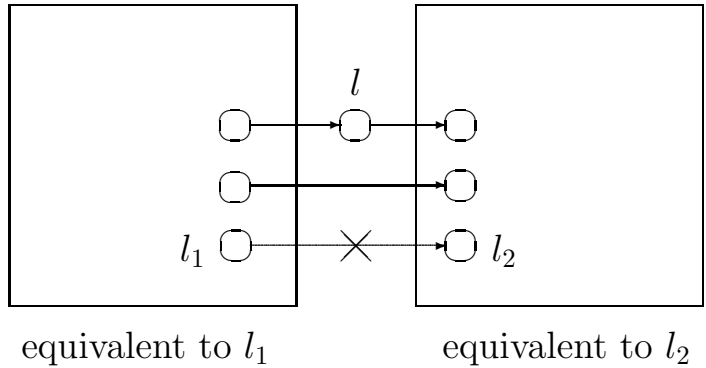

The idea is that, removing all the direct edges from the $l_1$ box to the $l_2$ box, then the edge $l_1 \rightarrow l_2$ is the only one that makes the boxes connected. This edge is therefore necessary in a subset $\Pi'$ of $\Pi$ that is equivalent to it. It is therefore in all I.E.S.'s of $\Pi'$, which are some I.E.S.'s of $\Pi$.

On the other hand, if there is a path that includes a node $l$ not in the boxes, then any I.E.S. includes a path from the $l_1$ box to $l$ and from $l$ to the $l_2$ box; otherwise, the node would not be either reachable from the $l_1$ box or the $l_2$ box would not be reachable from it. The presence of such path therefore makes $l_1 \rightarrow l_2$ always redundant.

**Lemma 19** *If $\Pi \not\models (l_1 \equiv l_2)$, then $l_1 \rightarrow l_2$ is in a I.E.S. if and only if all paths from $l_1$ to $l_2$ contain only literals that $\Pi$ makes equivalent either to $l_1$ or to $l_2$.*

*Proof.* Let us assume that all paths from $l_1$ to $l_2$ only contains literals that are equivalent to either $l_1$ or to $l_2$. Since $l_1$ and $l_2$ are not equivalent, any such a path can be written as $(l_1, \ldots, l_3, l_4, \ldots, l_2)$, where $\Pi$ makes literals $l_1, \ldots, l_3$ equivalent and literals $l_4, \ldots, l_2$ equivalent.

Let us now remove the edge $l_3 \rightarrow l_4$. By assumption, $l_4$ can still be reached from $l_3$ by first going to $l_1$, then to $l_2$, and then to $l_4$. As a result, one path from $l_1$ to $l_2$ has been removed while maintaining equivalence with $\Pi$.

Iterating this process over all paths from $l_1$ to $l_2$, we end up with a set of clauses whose only path from $l_1$ to $l_2$ is the single edge $l_1 \rightarrow l_2$. This clause is now irredundant: removing other redundant clauses, we obtain a I.E.S. with $l_1 \rightarrow l_2$.

Let us now assume the converse: there is a path from $l_1$ to $l_2$ that contains some literals that are not equivalent to $l_1$ nor to $l_2$. Such a path can be written as: $(l_1, \ldots, l_3, l_4, \ldots, l_5, l_6, \ldots, l_2)$, where $l_1, \ldots, l_3$ are equivalent to each other, as are $l_6, \ldots, l_2$, but are not equivalent to any literal in $l_4, \ldots, l_5$. We prove that no I.E.S. contains the edge $l_1 \rightarrow l_2$.

Let $\Pi'$ be an equivalent subset of $\Pi$. Being equivalent to $\Pi$, it must contain a number of cycles that make all literals of $l_1, \ldots, l_3$ equivalent to each other and all literals of $l_6, \ldots, l_2$ equivalent to each other. Note that $l_1 \rightarrow l_2$ cannot be in one of such cycles; otherwise, we would have a cycle joining both $l_1$ and $l_2$, which would prove that they are equivalent.

Since $\Pi \models l_1 \rightarrow l_4$ and $\Pi \models l_4 \rightarrow l_2$, the set $\Pi'$ must contain a path from $l_1$ to $l_4$ and a path from $l_4$ to $l_2$. If the first path includes $l_1 \rightarrow l_2$, then we would have a path from $l_2$ to $l_4$. Since $\Pi$ contains a path from $l_4$ to $l_2$, then $l_4$ and $l_2$ would be equivalent. For the same reason, the path from $l_4$ to $l_2$ do not contain the edge $l_1 \rightarrow l_2$.

As a result, we have proved that $\Pi'$ includes some sets of edges that do not contain $l_1 \rightarrow l_2$ but allow to conclude that $l_1, \ldots, l_3$ are equivalent to each other, that $l_6, \ldots, l_2$ are equivalent to each other, that $l_1$ implies $l_4$, and that $l_4$ implies $l_2$. The edge $l_1 \rightarrow l_2$ is therefore redundant. $\square$

This lemma implies that checking whether $l_1 \rightarrow l_2$ is in some I.E.S. is easy if $l_1$ is not equivalent to $l_2$. Indeed, the set of nodes in the paths from $l_1$ to $l_2$ can be found by intersecting the set of nodes that are reached from $l_1$ and that of the nodes that $l_2$ is reachable from. If this intersection contains a literal that is not equivalent to $l_1$ nor to $l_2$, then the edge $l_1 \rightarrow l_2$ is redundant.

## 6.3 Presence in an I.E.S.: Cyclic Inconsistent 2CNF Formulae

The problem of deciding the presence of a clause in an I.E.S. of an inconsistent 2CNF formula is NP-hard if the formula contains cycles. We use this simple preliminary lemma.

**Lemma 20** *The problem of checking the existence of a simple path from node $x$ to node $y$ in $G$ including a given edge is NP-complete.*

*Proof.* Membership is obvious. Hardness is proved by reduction from the problem path via a node: given a graph $G$ and three nodes $x$, $y$, and $m$, decide whether there exists a simple path from $x$ to $y$ including the node $m$.

The reduction is as follows: replace the node $m$ with two nodes $m_1$ and $m_2$ joined by an edge. For every edge $(n, m)$, add the edge $(n, m_1)$. For every edge $(m, n)$, add the edge $(m_2, n)$.

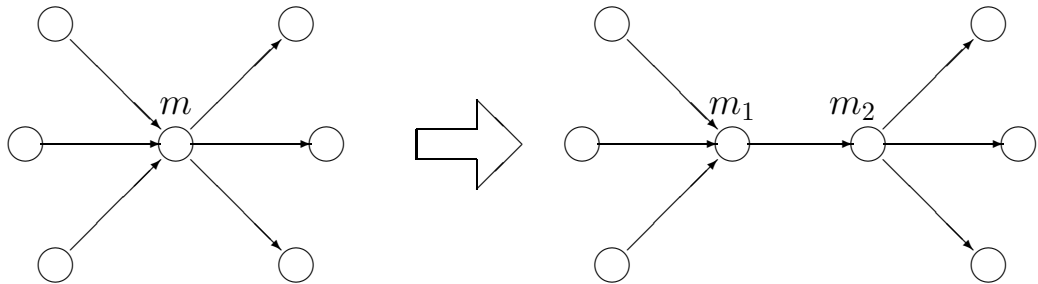

By construction, every simple path containing the node $m$ in the original graph contains an edge that is incoming to $m$ and is outgoing from $m$. This is possible if and only if we have a simple path in the new graph including the edeg $(m_1, m_2)$. □

We can now prove that the problem of presence of a clause in an I.E.S. is NP-complete if the formula is inconsistent and cyclic.

**Theorem 8** *Deciding whether a cyclic inconsistent 2CNF formula has an I.E.S. that contains the clause a given clause is NP-complete.*

*Proof.* Membership is obvious by gussing a subset of the formula and then checking whether it is an I.E.S. and contains the given clause. Hardness is proved by reduction from the problem of checking whether a graph contains a simple path from a node to another one that includes a given edge.

Given the instance of the original problem $(G, x, y, (a, b))$, in which we want to determine whether there exists a simple path from $x$ to $y$ including the edge $(a, b)$, we build a formula $\Pi$ as follows: for each node of the graph, we have a literal; for each edge $(n, m)$ of the graph, we have the clause $\neg n \to m$; finally, we add the following clauses, where $z$ and $w$ are new variables:

$$\{y \to z, y \to \neg z, \neg x \to w, \neg x \to \neg w\}$$

Graphically, Π is as follows.

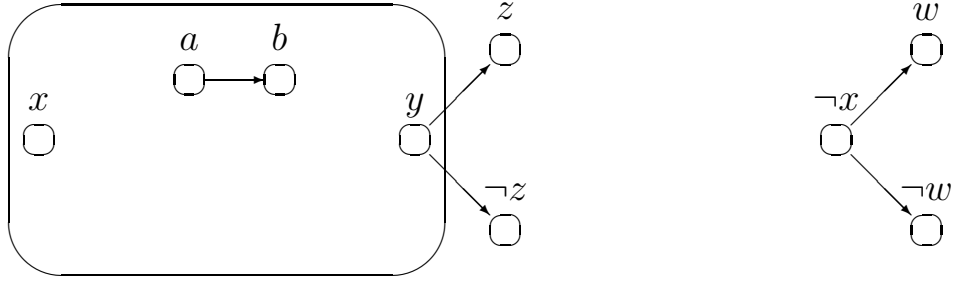

The part of the formula corresponding to the left part of the graph is satisfiable by setting all variables it contains to false. Indeed, this assigment makes all literals of the graph but $\neg z$ false; since all clauses contain at least one negated literal, they are all satisfied. By setting $x$ to true, the part of the formula corresponding to the graph on the right is satisfied.

Since both parts of Π are satisfiable, and $x$ is the only variable that is common to them, we have that $x$ is the variable such that $\Pi \cup \{x\} \models_{UP} \bot$ and $\Pi \cup \{\neg x\} \models_{UP} \bot$. Since $z$ is the only variable that occurs both positively and negatively in the first part of the formula, we have that $y$ must be reachable from $x$ in any I.E.S. of Π. As a result, there exists an acyclic path including the edge $(a, b)$ if and only if the edge $\neg a \vee b$ is in some I.E.S. of Π. □

## 6.4 Presence in an I.E.S.: Cyclic Consistent 2CNF Formulae Implying Literals

The I.E.S.'s of a consistent acyclic set can be compactly expressed as a number of independent choices. This makes the problem of checking the presence of a clause in an I.E.S. easy. In this section, we extend this polynomiality result to the case in which a literal of the clause to check is implied by the formula but the clause is not contained in any cycle of clauses. We also prove that the problem is instead NP-complete if the clause is contained in a cycle.

**Theorem 9** *The problem of deciding whether $\neg l_1 \vee l_2$ is in an I.E.S. of a consistent 2CNF Π is polynomial if $\Pi \models \neg l_1$ and $l_1$ is not reachable from $l_2$.*

*Proof.* The proof is similar to that of Lemma 15 with $CC_\Pi(l_1)$ in place of $l_1$ alone. The set $CC_\Pi(l_1)$ of literals that are in a cycle with $l_1$ can be determined in polynomial time. The clause $\neg l_1 \vee l_2$ is part of an I.E.S. of Π if and only if $\Pi \cup \{l_2\} \models_{UP} \bot$ or $\Pi \cup \{l_2\} \models_{UP} \neg l_3$, where $l_3 \in CC_\Pi(l)$.

This algorithm is correct because there is no path from $l_2$ to $l_1$. Therefore, there is no path from $l_2$ to any node of $CC_\Pi(l)$. As a result, the derivation from $l_2$ to $\bot$ or to $\neg l_3$ do not contain any literal in $CC_\Pi(l_1)$.

Let $\Pi'$ be an I.E.S. of Π. The derivations from $l_2$ to $\bot$ or to $\neg l_3$ still hold in $\Pi'$ and do not involve literals in $CC_\Pi(l_1)$. If $\bot$ is derivable from $l_2$, we can obtain an I.E.S. by by removing all clauses $\neg l \vee l'$ with $l \in CC_\Pi(l_1)$ and adding $\neg l_1 \vee l_2$ and a minimal number of clauses to make $l_2$ reachable from any literal in $CC_\Pi(l_1)$. This addition makes the negation of all literals of $CC_\Pi(l_1)$ entailed. This is therefore an I.E.S. because all clauses that have been removed contain the negation of a literal in $CC_\Pi(l_1)$.

A similar proof can be given for the case $\Pi \cup \{l_2\} \models_{UP} \neg l_3$. □

This theorem is about a clause $\neg l_1 \vee l_2$ that connects a node of $CC_\Pi(l_1)$ with a node outside it. Rewriting the clause as $l_1 \rightarrow l_2$ makes the idea more evident: the clause can be viewed as an edge that starts from $CC_\Pi(l)$, which is a set of literals all connected to each other, but ends outside $CC_\Pi(l_1)$. We now consider the case of a clause $\neg l_1 \vee l_2$ that is "internal" to $CC(l_1)$, that is, both $l_1$ and $l_2$ are in $CC(l_1)$. This conditions is equivalent to: there exists a cycle including both $l_1$ and $l_2$, which can be checked in polynomial time. The problem of presence of this clause in an I.E.S. is NP-complete.

**Theorem 10** *The problem of checking the presence of $\neg l_1 \vee l_2$ in some I.E.S.'s of a consistent 2CNF formula $\Pi$ such that $\Pi \models \neg l_1$ is NP-complete if $\Pi \cup \{l_2\} \models_{UP} l_1$.*

*Proof.* The proof is by reduction from the problem of deciding the existence of two vertex-disjoint paths in a directed graph, which is NP-complete [EIS76, FHW80] (the corresponding problem for undirected graphs is polynomial [RS04].) This is the problem of establishing whether a graph $G$ contains a path from node $s_1$ to node $t_2$ and a path from node $s_2$ to node $t_2$ and these two paths do not share nodes. We can assume that, from each node of the graph, either $t_1$ or $t_2$ is reachable. The formula we consider is the one corresponding to the following graph.

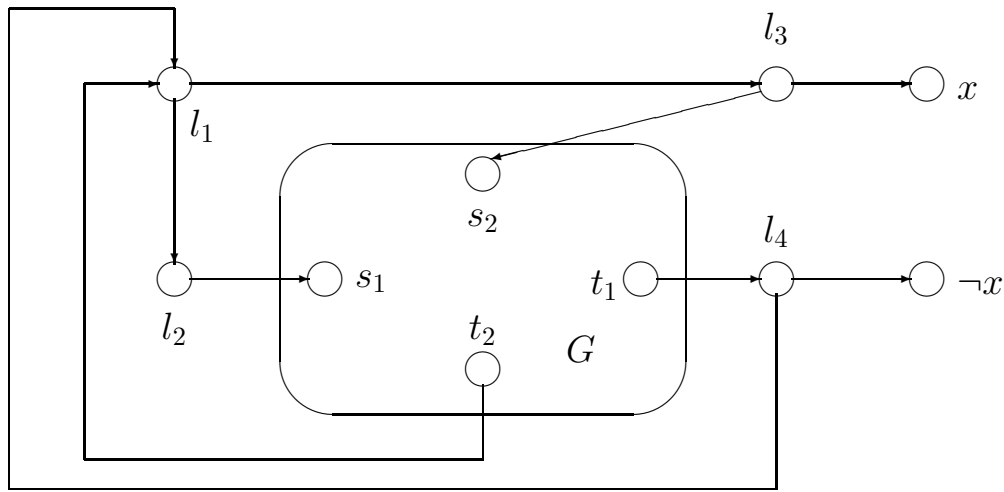

This formula implies the negation of all positive literals beside $x$. Indeed, the pair of nodes $x$ and $\neg x$ can be reached from any other node of the graph. The same property must therefore be true for all I.E.S.'s of this formula. Since $x$ and $\neg x$ form the only pair of opposite literals in the graph, the property holds only if both $l_3$ and $l_4$ are reachable from any other node besides $x$ and $\neg x$ in the graph corresponding to the I.E.S.

If $G$ has a pair of vertex-disjoint paths from $s_1$ to $t_1$ and from $s_2$ to $t_2$, respectively, an I.E.S. containing the clause corresponding to the edge $l_1 \to l_2$ is the subformula containing the clauses corresponding to the following edges:

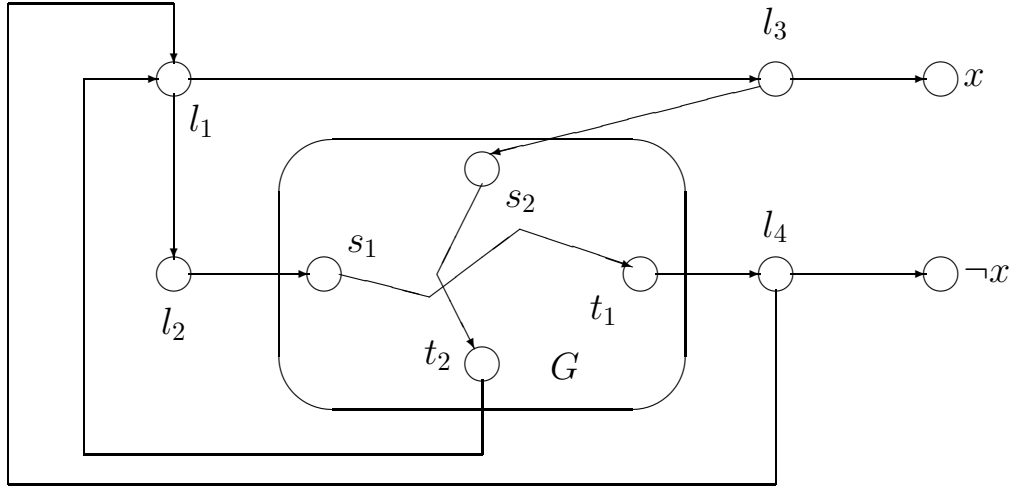

Additionally, the I.E.S. contains a number of edges making either $t_1$ or $t_2$ reachable from every node of $G$ not having already this property: this can be done by adding an edge $s \to t$ if either $t_1$ or $t_2$ is reachable from $t$ but not from $s$. The addition of these edges does not make $t_2$ reachable from $s_1$ or $t_1$ from $s_2$ because an edge $s \to t$ is only added if no other edge outgoing from $s$ is already in the I.E.S.

In this subgraph, both $l_3$ and $l_4$ are reachable from $l_1$ and $l_1$ is reachable from any other node of the graph. Since $l_4$ is only reachable from $l_1$ via the edge $l_1 \to l_2$, the corresponding clause is irredundant. Since $\neg l_1 \vee l_2$ is irredundant in this subformula, every I.E.S. of this subformula contains this clause by Property 1; since these I.E.S.'s are also I.E.S.'s of the original formula, we have that $\neg l_1 \vee l_2$ is contained in some I.E.S.'s of the original formula.

Let us now assume that $G$ contains no vertex-disjoint paths from $s_1$ to $t_1$ and from $s_2$ to $t_2$, respectively. We prove that $\neg l_1 \vee l_2$ is redundant in every I.E.S. of the formula. Since $x$ is the only variable that occur both direct and negated in the graph, in every I.E.S. of $\Pi$ both $l_3$ and $l_4$ are reachable from any other node of the graph besides $x$ and $\neg x$. The following are the edges that are necessarily contained in any I.E.S. of $\Pi$ because either they are the only outgoing edges of a node or they are the only edges that are incoming to $l_3$ or $l_4$.

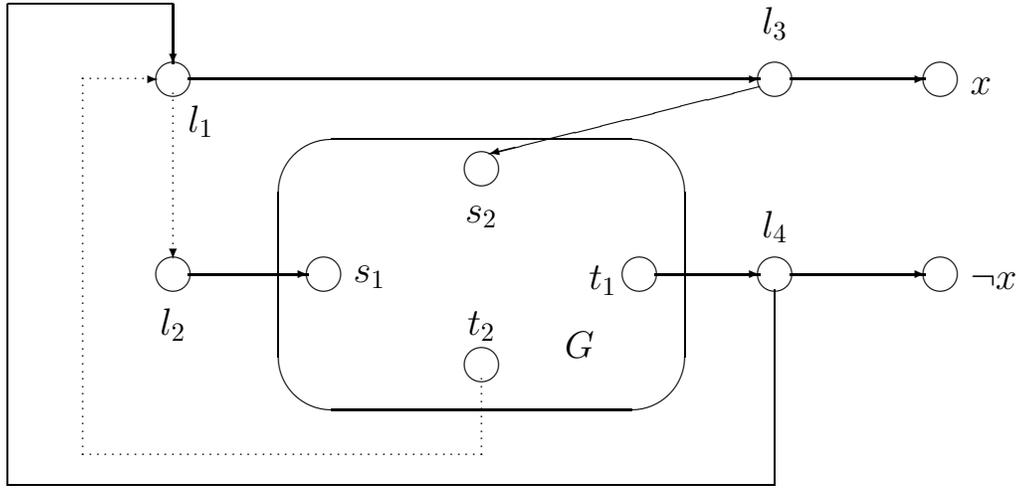

Since $s_2$ is connected to nodes outside $G$, it is either connected to $t_1$ or to $t_2$. If $t_1$ is reachable from $s_2$, then both $l_3$ and $l_4$ are reachable from $l_1$, and the edge $l_1 \to l_2$ is redundant. We can therefore consider only the case in which $s_2$ is connected to $t_2$ but not to $t_1$.

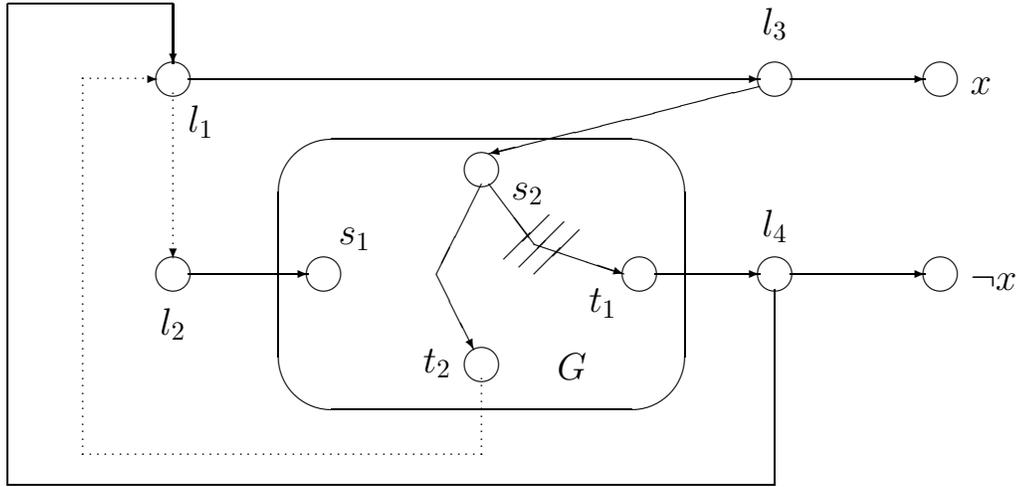

In this graph, $t_1$ is not reachable from $s_2$ using only edges inside $G$. The other nodes of the graph can therefore be connected to $l_4$ only via a path from $s_1$ to $t_1$. By assumption, however, every path from $s_1$ to $t_1$ shares a node with the path from $s_2$ to $t_2$, which makes $t_1$ reachable from $s_2$, contradicting the assumption. □

## 6.5 Presence in an I.E.S.: Cyclic Consistent 2CNF Formulae not Implying Literals

The problem of presence of a clause in an I.E.S. of a 2CNF formula is polynomial if the formula is acyclic and consistent. We now show that the same problem is NP-complete if the formula is still consistent but cyclic. First of all, if a formula is consistent and does not entail literals, then every clause $l_1 \vee l_2$ can be made irredundant by the following lemma.

**Lemma 21** *If $\Pi$ does not entail $\neg l_1$ nor $\neg l_2$ and contains the clauses $l_1 \vee w$ and $\neg w \vee l_2$, where $w$ is a variable appearing only in these two clauses, then $\Pi$ is equivalent to $\Pi \cup \{l_1 \vee l_2\}$, and the two clauses $l_1 \vee w$ and $\neg w \vee l_2$ are irredundant in $\Pi$.*

*Proof.* Equivalence is due to the fact that $l_1 \vee l_2$ is obtained from $l_1 \vee w$ and $\neg w \vee l_2$ by resolution. Irredundancy is due to the fact that $\Pi \cup \{\neg l_1\} \models_{UP} w$ in $\Pi$ because $\Pi$ is consistent and does not entail literals. Since $l_1 \vee w$ is the only clause containing $w$ positively, this clause is necessary. For the same reason, $\neg w \vee l_2$ is necessary. $\square$

If a clause is necessary in a formula, it is containing in all its I.E.S.'s. Whenever we have a formula $\Pi$ that is consistent and does not entail literals, and we want a clause $l_1 \vee l_2$ to be contained in all its I.E.S.'s, we can then replace it with $l_1 \vee w$ and $\neg w \vee l_2$. The original clause is still implied by these two new ones by resolution, but the new clauses are necessary. In order to keep proofs simple, we use the following notation.

NOTATION:    $l_1 \overset{\circ}{\vee} l_2$ means $\{l_1 \vee w, \neg w \vee l_2\}$ where $w$ is a new variable

We use this notation because the clauses $\{l_1 \vee w, \neg w \vee l_2\}$ actually represent the single clause $l_1 \vee l_2$ since $w$ is not used anywhere else. The circle over the symbol $\vee$ reminds us that these two clauses cannot be removed from a formula without changing its semantics. We can now prove that deciding the presence of a clause in an I.E.S. of a formula is NP-complete.

**Theorem 11** *Deciding whether a clause is in an I.E.S. of a cyclic and consistent 2CNF formula $\Pi$ is NP-hard even if no single literal is implied by $\Pi$ and $\Pi$ makes all literals equivalent.*

*Proof.* We show a proof of hardness from 3sat. The set of clauses generated by this particular reduction is such that all clauses are of the form $\neg l \vee l'$. Such formulae can be represented by their induced graphs. In this proof, we use $l \rightarrow l'$ to denote $\neg l \vee l'$ and $l \overset{\circ}{\rightarrow} l'$ to denote $\neg l \overset{\circ}{\vee} l'$. We also use $l \Rightarrow l'$ to denote the reachability of $l'$ from $l$ and $l \Leftrightarrow l'$ to denote that $l$ and $l'$ can be reached from each other.

Given a set of clauses $\Gamma = \{\gamma_1, \ldots, \gamma_m\}$, we generate a formula $\Pi$ and one of its clauses $l_1 \rightarrow l_2$ in such a way $l_1 \rightarrow l_2$ in is in some I.E.S.'s of $\Pi$ if and only if $\Gamma$ is satisfiable. The graph corresponding to $\Pi$ is strongly connected. In particular, truth assignments on $\Gamma$ correspond to subsets of $\Pi$ in which $l_2 \Rightarrow n \Rightarrow l_1$ for every node $n$. Therefore, the graph is strongly connected if and only if $l_1 \Rightarrow l_2$. The truth assigment satisfies $\Gamma$ if and only if $l_1 \Rightarrow l_2$ does not hold in the corresponding subformula, thus making the clause $l_1 \rightarrow l_2$ necessary.

From now on, we consider $\Pi$ as a graph. The graph corresponding to $\Gamma$ is as follows: for each variable $x_i$, we have three nodes $x_i$, $x_i^+$, and $x_i^-$. For each clause $\gamma_j$ we have a node $c_j$. The edges of $\Pi$ are the following ones:

1. Nodes forming a strongly connected component containing $l_1$:

(a) $l_1 \overset{\circ}{\to} x_i$;

(b) $x_i \to x_i^+$ and $x_i \to x_i^-$;

(c) $x_i^+ \overset{\circ}{\to} l_1$ and $x_i^- \overset{\circ}{\to} l_1$;

2. Nodes forming a strongly connected component containing $l_2$:

   (a) $l_2 \overset{\circ}{\to} x_i^+$ and $l_2 \overset{\circ}{\to} x_i^-$;

   (b) $x_i^+ \to c_j$ if $x_i \in \gamma_j$;

   (c) $x_i^- \to c_j$ if $\neg x_i \in \gamma_j$.

   (d) $\gamma_j \overset{\circ}{\to} l_2$;

This graph is strongly connected. This must also be true for every graph representing an I.E.S. of $\Pi$. Graphically, the clause $\gamma_1 = x_1 \vee \neg x_2$ is represented as in Figure 1. The nodes $l_1$ and $l_2$ have been omitted to keep the figure simple: if a node is missing at the left of an arrow, it is $l_1$; if it is at the right, it is $l_2$. Arrows marked with a circle cannot be removed while looking for an I.E.S. of this formula.

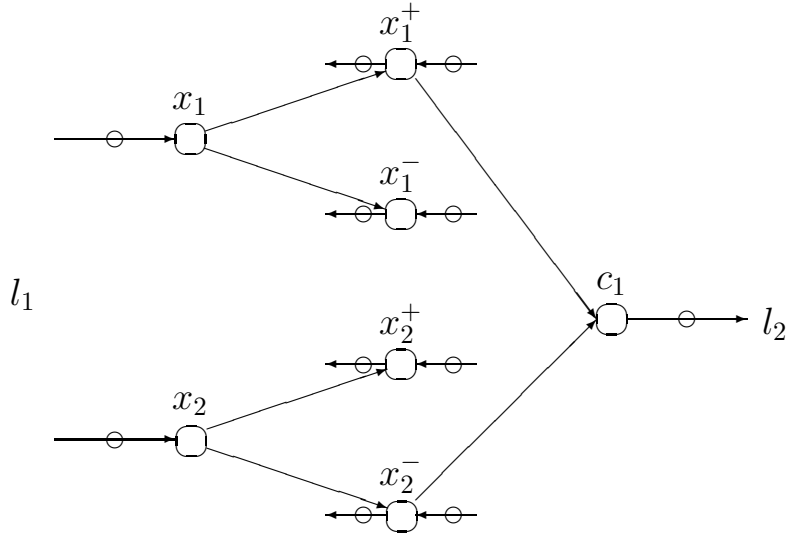

Figure 1: The subgraph corresponding to $\gamma_1 = x_1 \vee \neg x_2$.

Every truth assigment on $\Gamma$ can be associated to a set of edges to remove from $\Pi$ to the aim of obtaining an I.E.S. In particular, if $\Pi$ is satisfiable we can remove some of the edges in such a way $l_2 \Rightarrow n \Rightarrow l_1$ still hold for every node $n$, but $l_1 \Rightarrow l_2$ does not. This way $l_1 \to l_2$ is necessary to make this graph strongly connected and therefore equivalent to the original one.

Let us assume that $\Gamma$ is satisfiable, and let $M$ be one of its models. We build a subset of $\Pi$ by removing the following edges:

1. if $x_i$ is positive in $M$, remove the edge $x_i \to x_i^+$ and all edges from $x_i^-$ to a node $c_j$;

2. if $x_i$ is negative in $M$, do the other way around.

The following figure shows the edges that are removed from the formula above if $M$ assigns true to both $x_1$ and $x_2$.

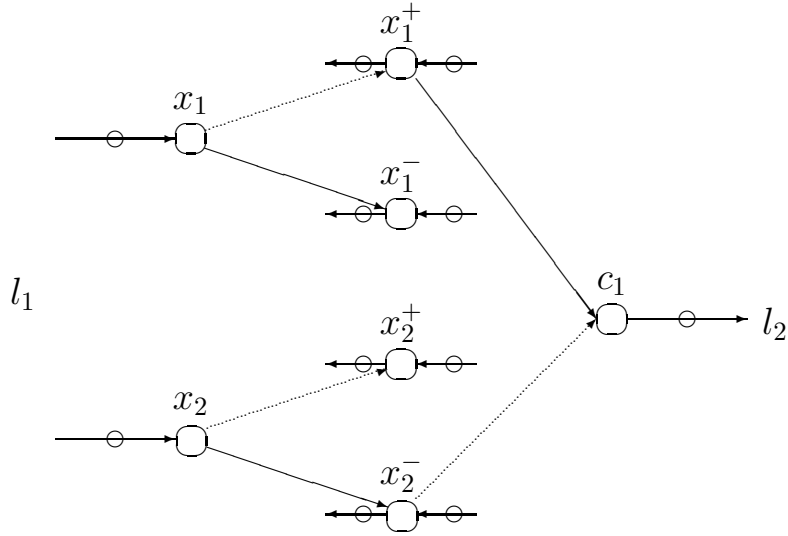

In this subgraph, $l_2 \Rightarrow n \Rightarrow l_1$ holds for all nodes. For nodes $x_i^+$ and $x_i^-$ there are edges from $l_2$ to them and from them to $l_1$. The cycle $l_1 \Rightarrow x_i \Rightarrow x_i^+ \Rightarrow l_1$ or the cycle with $x_i^-$ replacing $x_i^+$ makes all nodes $x_i$ in the same strongly connected component with $l_1$ if $x_i$ is negative in $M$. A node $c_j$ is in the same strongly connected component with $l_2$ thanks to a cycle $l_2 \to x_i^+ \to c_j \to l_2$ with $x_i$ is positive in $M$ and in $c_j$, or the similar cycle if the literal of $c_j$ that is true in $M$ is a negative one.

Since $l_2 \Rightarrow n \Rightarrow l_1$ holds for all nodes, the addition of the edge $l_1 \to l_2$ makes all nodes reachable from each other, making this formula equivalent to the original one. On the other hand, $l_1 \Rightarrow l_2$ is not true in this graph: indeed, all paths $l_1 \to x_i \to x_i^+ \to c_j \to l_2$ have been broken because either the edge $x_i \to x_i^+$ or the edge $x_i^+ \to c_j$ have been removed, and the same for the similar path containing $x_i^-$ in place of $x_i^+$.

Let us now assume that $\Pi$ has an I.E.S. $\Pi'$ containing the clause $l_1 \to l_2$, and show a truth assigment satisfying all clauses of $\Gamma$. Since the graph of $\Pi$ is strongly connected, the same holds for $\Pi'$. Since removing $l_1 \to l_2$ makes the graph of $\Pi'$ not strongly connected, we have that:

1. $l_2$ is not reachable from $l_1$ in the graph corresponding to $\Pi' \backslash \{l_1 \to l_2\}$;

2. every node is reachable from either $l_1$ and $l_2$ and reaches either $l_1$ and $l_2$ in the graph corresponding to $\Pi' \backslash \{l_1 \to l_2\}$: otherwise, the addition of $l_1 \to l_2$ would not make the graph strongly connected.

Since $x_i$ is reachable from $l_1$, it must reach either $l_1$ or $l_2$; however, it cannot reach $l_2$ as otherwise we would have $l_1 \Rightarrow l_2$. As a result, for every $x_i$, either $x_i \to x_i^+$ or $x_i \to x_i^-$ is in $\Pi'$. In other words, it cannot be that both edges are not in $\Pi'$ for the same $x_i$. As a result, the following is a partial model (a consistent set of literals).

$$M = \{x_i \mid x_i \to x_i^- \notin \Pi'\} \cup \{\neg x_i \mid x_i \to x_i^+ \notin \Pi'\}$$

We show that all clauses of $\Gamma$ are satisfied by $M$. Let $\gamma_j$ be a clause. Since the edge $c_j \to l_2$ is not removable, $l_1 \Rightarrow c_j$ cannot hold, as otherwise $l_2$ would be reachable from $l_1$. As a result, $l_2 \Rightarrow c_j$ must hold. This implies that there exists an index $i$ such that either $x_i \in \gamma_j$ and $x_i^+ \to c_j \in \Pi'$ or $\neg x_i \in \gamma$ and $x_i^- \to c_j \in \Pi'$. In the first case, $x_i \to x_i^+ \notin \Pi'$ as this edge would create a path from $l_1$ to $l_2$. This however implies that $x_i$ is set to true by $M$. Since $x_i \in \gamma_j$, the clause $\gamma_j$ is satisfied by $M$. The case $\neg x_i \in \gamma_j$ and $x_i^- \to c_j \in \Pi'$ is similar. □

# 7 Horn Formulae

When considering the complexity of redundancy for Horn formulae, the following problems are clearly polynomial.

1. checking redundancy;

2. a set is an I.E.S.;

3. a clause is in all I.E.S.'s;

4. uniqueness.

The problem of size is easily proved to be NP-complete: indeed, the corresponding proof for the case of consistent acyclic 2CNF formulae not implying single literals uses clauses corresponding to the edges of a graph whose nodes are all positive literals. These clauses are therefore all in the form $\neg x \vee y$, that is, they are binary Horn clauses.

We show that the problem of size is NP-complete for inconsistent Horn formulae.

**Theorem 12** *Deciding whether a Horn formula $\Pi$ has an equivalent subset of size $k$ is NP-complete.*

*Proof.* Membership if obvious. Hardness is proved by reduction from vertex cover. Let $G$ be a graph. We build the following set of Horn clauses: for each node $i$ we have a unit clause $x_i$; for each edge $z = (i, j)$ we have two clauses $x_i \to a_z$ and $x_j \to a_z$; finally, we have the clause $\neg a_1 \vee \cdots \vee \neg a_m$.

Inconsistent subsets of this formula are composed of $\neg a_1 \vee \cdots \vee \neg a_m$, plus a pair $x_i$ and $x_i \to a_z$ for each edge $z$. This means that we have exactly one clause $x_i \to a_z$ for each edge of the graph. Moreover, for each edge we have to include a unit clauses corresponding to one of its incident nodes. Therefore, minimal inconsistent subsets are in one-to-one correspondence with vertex covers of the original graph. Namely, $G$ has a vertex cover of $k$ nodes if and only if the formula has an inconsistent subset of $m + 1 + k$ clauses, where $m$ is the number of edges of $G$. □

# A  Notations

$$
\begin{aligned}
C_\Pi(l) &= \{l' \mid \neg l \vee l' \in \Pi\} \\
D_\Pi(l) &= \{\gamma \in \Pi \mid \gamma = \neg l \vee l'\} \\
R_\Pi(l) &= \Pi \backslash D_\Pi(l) \\
M_\Pi(l) &= \{l' \in C_\Pi(l) \mid \nexists l'' \in C_\Pi(l) \text{ such that } R_\Pi(l) \cup \{l''\} \models_{UP} l'\} \\
S_\Pi(l) &= \{l' \in C_\Pi(l) \mid R_\Pi(l) \cup \{l'\} \models_{UP} \bot\} \\
P_\Pi(l) &= \{(l_1, l_2) \mid l_1, l_2 \in C_\Pi(l) \text{ and } R_\Pi(l) \cup \{l_1\} \models_{UP} \neg l_2\} \backslash S_\Pi(l) \\
CC_\Pi(l) &= \{l' \mid l \text{ and } l' \text{ are in a cycle}\} \\
JC_\Pi(l) &= \{l' \mid \neg l'' \vee l' \in \Pi \text{ and } l'' \in CC_\Pi(l)\} \backslash CC_\Pi(l) \\
LC_\Pi(l) &= \{l' \in JC_\Pi(l) \mid \Pi \cup \{l\} \models_{UP} l'\}
\end{aligned}
$$

# Contents

# References

[ADS86]   G. Ausiello, A. D'Atri, and D. Saccà. Minimal representation of directed hypergraphs. *SIAM Journal on Computing*, 15(2):418–431, 1986.

[Bru03]   R. Bruni. Approximating minimal unsatisfiable subformulae by means of adaptive core search. *Discrete Applied Mathematics*, 130(2):85–100, 2003.

[BZ05]    H. Büning and X. Zhao. Extension and equivalence problems for clause minimal formulae. *Annals of Mathematics and Artificial Intelligence*, 43(1):295–306, 2005.

[EIS76]   S. Even, A. Itai, and A. Shamir. On the complexity of timetable and multicommodity flow problems. *SIAM Journal on Computing*, 5(4):691–703, 1976.

[FHW80]   S. Fortune, J. Hopcroft, and J. Wyllie. The directed subgraph homeomorphism problem. *Theoretical Computer Science*, 10(2):111–121, 1980.

[FKS02]   H. Fleischner, O. Kullmann, and S. Szeider. Polynomial-time recognition of minimal unsatisfiable formulas with fixed clause-variable difference. *Theoretical Computer Science*, 289:503–516, 2002.

[GF93]    G. Gottlob and C. G. Fermüller. Removing redundancy from a clause. *Artificial Intelligence*, 61:263–289, 1993.

[Gin88]   A. Ginsberg. Knowledge base reduction: A new approach to checking knowledge bases for inconsistency & redundancy. In *Proceedings of the Seventh National Conference on Artificial Intelligence (AAAI'88)*, pages 585–589, 1988.

[HK93]    P. Hammer and A. Kogan. Optimal compression of propositional Horn knowledge bases: Complexity and approximation. *Artificial Intelligence*, 64(1):131–145, 1993.

[HW97]    E. Hemaspaandra and G. Wechsung. The minimization problem for Boolean formulas. In *Proceedings of the Thirtyeighth Annual Symposium on the Foundations of Computer Science (FOCS'97)*, pages 575–584, 1997.

[KRF95]   S. Khuller, B. Raghavachari, and M. Fellows. Approximating the minimum equivalent digraph. *SIAM Journal on Computing*, 24(4):859–872, 1995.

[Lib]     P. Liberatore. Redundancy in logic III: Non-classical logics. Manuscript.

[Lib05]   P. Liberatore. Redundancy in logic I: CNF propositional formulae. *Artificial Intelligence*, 163(2):203–232, 2005.

[LP84]    A. LaPaugh and C. Papadimitriou. The even-path problem for graphs and digraphs. *Networks*, 14:507–513, 1984.

[Mai80]   D. Maier. Minimum covers in relational database model. *Journal of the ACM*, 27(4):664–674, 1980.

[MS72]    A. Meyer and L. Stockmeyer. The equivalence problem for regular expressions with squaring requires exponential space. In *Proceedings of the Thirteenth Annual Symposium on Switching and Automata Theory (FOCS'72)*, pages 125–129, 1972.

[PW88]    C. Papadimitriou and D. Wolfe. The complexity of facets resolved. *Journal of Computer and System Sciences*, 37:2–13, 1988.

[RS04]    N. Robertson and P. Seymour. Graph minors. XX. Wagner's conjecture. *Journal of Combinatorial Theory, Series B*, 92:325–357, 2004.

[Sah74]    S. Sahni. Computationally related problems. *SIAM Journal on Computing*, 3(4):262–279, 1974.

[SS97]    J. Schmolze and W. Snyder. Detecting redundant production rules. In *Proceedings of the Fourteenth National Conference on Artificial Intelligence (AAAI'97)*, pages 417–423, 1997.

[Uma98]    C. Umans. The minimum equivalent DNF problem and shortest implicants. In *Proceedings of the Thirtynineth Annual Symposium on the Foundations of Computer Science (FOCS'98)*, pages 556–563, 1998.